%% file: main.tex
\documentclass[letterpaper]{article} 
\usepackage{aaai25}  
\usepackage{times}  
\usepackage{helvet}  
\usepackage{courier}  
\usepackage[hyphens]{url}  
\usepackage{graphicx} 
\usepackage{natbib}  
\usepackage{caption} 
\usepackage{amsmath}
\usepackage{amssymb}
\usepackage{mathtools}
\usepackage{amsthm}
\usepackage{subfigure}
\usepackage{fancyhdr}
\usepackage{xcolor}
\usepackage{times}
\usepackage{booktabs}

\usepackage{float}
\usepackage{enumitem}
\usepackage[export]{adjustbox}

\theoremstyle{plain}
\newtheorem{theorem}{Theorem}[section]

\theoremstyle{definition}

\theoremstyle{remark}
\newtheorem{remark}[theorem]{Remark}

\input{preample_shared}

\newcommand{\dt}{{\Delta t}}
\newif\iflong
\longfalse

\begin{document}

\captionsetup[table]{position=bottom}

\title{Semi-Implicit Neural Ordinary Differential Equations}

\author{
Hong Zhang \textsuperscript{\rm 1},
Ying Liu \textsuperscript{\rm 2},
Romit Maulik \textsuperscript{\rm 1,\rm 3}
}
\affiliations {
    \textsuperscript{\rm 1}Argonne National Laboratory\\
    \textsuperscript{\rm 2}University of Iowa\\
    \textsuperscript{\rm 3}Pennsylvania State University\\
    hongzhang@anl.gov, ying-liu-1@uiowa.edu, rmm7011@psu.edu
}

\maketitle

\section*{Abstract}
Classical neural ODEs trained with explicit methods are intrinsically limited by stability,
crippling their efficiency and robustness for stiff learning problems that are common in graph learning and scientific machine learning. 
We present a semi-implicit neural ODE approach that exploits the partitionable structure of the underlying dynamics.
Our technique leads to an implicit neural network with significant computational advantages over existing approaches because of enhanced stability and efficient linear solves during time integration.
We show that our approach outperforms existing approaches on a variety of applications including graph classification and learning complex dynamical systems.  
We also demonstrate that our approach can train challenging neural ODEs where both explicit methods and fully implicit methods are intractable.

\section{Introduction}

Implicit neural networks (NNs) \cite{Winston2020, Bai2019} have emerged as a new paradigm for designing infinite-depth networks. The traditional neural network consists of an \emph{explicit} stack of layers. In contrast, an implicit neural network
represents an implicit model that often involves a nonlinear system, such as neural ordinary differential equations (ODEs) \cite{Chen2018,Dupont2019,Massaroli2020,kidger2022neural} and equilibrium models \cite{Winston2020, Bai2019}.
In order to solve the nonlinear system, an iterative procedure is often needed. For example, neural ODEs require a time integrator to obtain the solution at desired time points. Deep equilibrium (DEQ) models \cite{Bai2019} solve an implicit system with fixed-point iteration. Differentiating through the iterative procedure with direct backpropagation can be intractable, however, due to the memory consumption that increases linearly.
Key to the success of existing implicit NNs is using high-level adjoint methods for gradient calculation. 
The adjoint methods can reduce the memory footprint significantly for both equilibrium models \cite{Winston2020} and ODE models \cite{Chen2018,zhuang2020, Zhang2022pnode}. 

Despite the advancement in gradient calculation approaches, however, implicit NNs are still limited by the bottleneck created by computationally expensive algorithms for solving the \emph{linear or nonlinear system}.
New methods have been developed to circumvent the need to solve the original linear systems. The Jacobian-free backpropagation (JFB) method in \cite{Fung2022} exploits a preconditioning technique. Proximal implicit neural ODEs \cite{baker2022proximal} approximate the solution of the linear system with an optimization algorithm. However, all these methods are rooted in low-order iterative procedures, which eases differentiation but suffers from suboptimal convergence speed and instability.
Traditional linear solvers, including those highly optimized for AI accelerators such as GPUs, have favorable convergence properties but 
are not widely used in machine learning
due to training efficiency constraints.

For neural ODEs, an additional bottleneck comes from the \emph{numerical stability}.
Explicit solvers, especially those based on Runge-Kutta (RK) methods, have been the default choice for the training of neural ODEs.
These methods are only \emph{conditionally} stable, requiring small time steps to maintain stability.
The step size can be extremely small for \emph{stiff} ODEs.
Stiffness may result from dramatically different time scales involved in the dynamics that the ODE models, or from the spatial discretization of partial differential equations (PDEs) on a grid.
For example, the Graph Neural Diffusion (GRAND) model, recently proposed for building continuous-depth graph NNs (GNNs) \cite{Xhonneux2020} using graph neural ODEs \cite{poli2019graph}, results in stiff diffusion equations.
With explicit methods, one has to use a step size as small as $0.005$ to main stability, leading to $43,908$ NN evaluations per epoch. See Sec.~\ref{sec:exp} for details on the experiments on GRAND and more examples.

This work makes the following contributions:
\begin{itemize}[noitemsep]
\item We propose a \emph{semi-implicit} neural ODE (SINODE), a new type of implicit neural network, based on a linear-nonlinear \emph{partitioning} of the system.
The partitioned system is integrated with implicit-explicit (IMEX) methods, providing significant improvement in computational efficiency and stability over traditional neural ODEs.
\item We develop a \emph{differentiable} IMEX ODE solver as the backbone of SINODE. 
The solver provides reverse-accurate gradients through a \emph{discrete adjoint} approach, avoiding the cost of backpropagation through the ODE solver; it also enables state-of-the-art linear solution techniques when solving the implicit functions for mini-batch training of SINODE.
\item  We demonstrate the efficacy and efficiency of SINODE with graph learning tasks and two time-series prediction problems that are inherently \emph{nonlinear}, \emph{stiff}, or \emph{chaotic}. 
The results also show that off-the-shelf solvers can be adapted to machine learning models to provide \emph{stability} and \emph{efficiency}.
\end{itemize}

\section{Preliminaries}

\textbf{Adjoints and Automatic Differentiation} Reverse-mode automatic differentiation, as widely used in scientific computing and machine learning, is also an adjoint method implemented at a low level.
The adjoint methods used for implicit networks are derived at a high level of abstraction. We will take the DEQ model and the neural ODE model as examples.
In a continuous view, the forward pass of the DEQ model can be considered as a root-finding procedure for a nonlinear function, and the loss depends on the root:
\begin{equation}
z = \f(z, u; p), \quad \text{Loss } \ell = \Phi (z), 
\end{equation}
where $z$ is the equilibrium state for input $u$ and $\f$ stands for the NN parameterized by $p$.
The adjoint equation of the nonlinear problem \cite{Bai2019} is a linear equation defined as
\begin{equation}
\left( \I - \frac{\partial \f(z, u; p)}{\partial z} \right)^T \blambda =  \left(\frac{d \ell }{d z}\right)^T,
\end{equation}
where the derivative $\frac{d \ell }{d z}$ is a row vector and the adjoint variable $\blambda$ is a column vector.
The loss gradient (a column vector) can be calculated with
\begin{equation} 
\nabla_p \ell =  \left(\frac{\partial \f(z,u; p)}{\partial p} \right)^T \blambda.
\end{equation}
The adjoint model requires storing only $z$ and $u$ during training; therefore, the memory cost is independent of the number of iterations in the forward pass.

For an autonomous neural ODE
\begin{equation}
\frac{du(t)}{dt} = \f(u(t); p), \quad t \in [0, T],  \quad \text{Loss } \ell = \Phi (u(T)), 
\label{eqn:ode}
\end{equation}
its continuous adjoint \cite{Kidger2020HeyTN} is a new ODE integrated backward from time $T$ to $0$:
\begin{equation}
\frac{d \blambda(t)}{dt} = -\left(\frac{\partial \f(u(t); p)}{\partial u(t)}\right)^T \blambda(t).
\label{eqn:adjoint_ode}
\end{equation}
The loss gradient is computed with
\begin{equation} 
\nabla_p \ell = \int_0^T \left(\frac{\partial \f(u(t); p)}{\partial p} \right)^T \blambda(t) dt.
\end{equation}
For notational brevity, throughout this paper we omit $p$ in all parameterized functions and drop the arguments $(\cdot)$ for functions and variables such as $\f$, $u$, and $\blambda$ when there is no ambiguity.

\textbf{Reverse-accurate Adjoint Approaches}
In a discrete adjoint approach, the adjoint is derived based on
the discretized version of the continuous equations \eqref{eqn:ode}.
Without loss of generality, we assume \eqref{eqn:ode} is discretized with a time-stepping algorithm, denoted by an operator $\N$ that propagates the solution from one time step to another:
\begin{equation}
    u_{n+1} = \N (u_n), \quad n = 0, 1, \dots, N.
    \label{eq:ts_alg}
\end{equation}
The adjoint model of this time-stepping algorithm is
\begin{equation}
  \begin{aligned}
  \blambda_{n} &=  \left(\frac{\partial \N}{\partial u}(u_n)\right)^T \blambda_{n+1}, \\
  \bmu_{n} & =  \left(\frac{\partial \N}{\partial p}(u_n)\right)^T \blambda_{n} + \bmu_{n+1}, \, n= N-1, \dots, 0,
  \end{aligned}
\label{eqn:disadjoint}
\end{equation}
with terminal condition $\blambda_{N} = \left(\frac{\partial \ell}{\partial u_N}\right)^T, \quad \bmu_{N} = \left(\frac{\partial \ell}{\partial p}\right)^T.$

The adjoint variables $\blambda$ and $\bmu$ denote the loss gradient with respect to the state $u$ and the NN parameters $p$, respectively.
For a practical implementation, \eqref{eqn:disadjoint} is typically derived in a case-by-case manner for different time-stepping algorithms \cite{Zhang2014}.
The discrete adjoint model is conceptually equivalent to applying automatic differentiation to each iteration of the time-stepping procedure, therefore yielding the equally accurate gradient as obtained with direct backpropagation. 
Since the same trajectory (sequence of time steps) is used for the forward pass and the backward pass, one only needs to control step sizes during the forward pass when using adaptive time steps with the discrete adjoint approach.

\section{The Partitioned Model}

We consider an autonomous neural ODE framework with an additively partitioned right-hand side:
\begin{equation}
\label{eq:PETSC:ARK:partitioning}
\frac{d u}{dt} = \bG(u) + \bH(u) \,,
\end{equation}
where $u(t)\in \R^{d}$ is the state,  $t$ is the time, and $\bG: \R^{d} \rightarrow \R^{d}$ and $\bH: \R^{d} \rightarrow \R^{d}$ are NNs, for example feed-forward NNs. Similar to conventional neural ODE, our framework takes an input $u(t_0)$, learns the representation of the dynamics, and predicts the output $u(t_N)$ or a sequence of outputs ${u_{t_i}, t=1,\dots,N}$ by solving \eqref{eq:PETSC:ARK:partitioning} from the starting time $t_0$ to the ending time $t_N$ with a numerical ODE solver. While this framework is general and applicable for a wide range of applications, we are interested in a specific model with a nonlinear part and a linear part. 
For a purely linear neural ODE, which is rare in practice, we can use a partitioned model containing two linear parts.
Without loss of generality, we assume $\bH$ is linear, and thus \eqref{eq:PETSC:ARK:partitioning} can be rewritten to

\begin{equation}
\label{eqn:nonlinear-linear-partitioning}
\frac{d u}{dt} =  \underset{\text{Nonlinear}}{\underbrace{\bG(u)}} +
\underset{\text{Linear}}{\underbrace{ \vphantom{\bG(u)} \J u}}\,.
\end{equation}
The reason for this choice will become clear later.

\section{Training Algorithms}
To train the partitioned NN, we adopt a semi-implicit approach to integrate \eqref{eq:PETSC:ARK:partitioning} in a forward pass and compute the gradients with a discrete adjoint approach in a backward pass.
The algorithm for each pass will be described below.

\subsection{Forward Pass} \label{sec:fwd}

In the forward pass, we solve the partitioned ODE \eqref{eq:PETSC:ARK:partitioning} using implicit-explicit RK  (IMEX-RK) methods. 
In particular, the right-hand side $\bG$ is treated explicitly while $\bH$ is treated implicitly.
At each time step, the solution is advanced from $t_n$ to $t_{n+1}$ using
\begin{small}
\begin{equation}
\begin{aligned}
U^{(i)} &= u_{n} + \dt \sum_{j=1}^{i-1} a_{ij} \bG^{(j)} +
\dt \sum_{j=1}^{i} \tilde{a}_{ij} \bH^{(j)},~ i=1,\dots,s\\
u_{n+1} &= u_{n} + \dt \sum_{j=1}^{s} b_{j} \bG^{(j)} +
\dt \sum_{j=1}^{s} \tilde{b}_{j} \bH^{(j)}\,,
\end{aligned}
\label{eq:ARK:ODE:compl}
\end{equation}
\end{small}
where $U^{(i)}$ is the stage value, $\dt$ is the step size, $A=\{a_{ij}\}$ is strictly lower triangular,
$\tilde{A}=\{\tilde{a}_{ij}\}$ is lower triangular
and can have zeros on the diagonal (these correspond to explicit
stages), and  $\circ^{(i)}$ denotes stage index $i$. $\bG^{(j)}$ and $\bH^{(j)}$ are shorthand for $\bG(U^{(j)})$ and $\bH(U^{(j)})$.
The IMEX-RK scheme can be represented by the Butcher tableau consisting of $(A=\{a_{ij}\},b,c)$ for the explicit part and $(\tilde{A}=\{\tilde{a}_{ij}\},\tilde{b},\tilde{c})$ for the implicit part. See the appendix for the coefficients of the IMEX methods used in this paper.
The IMEX-RK methods are well known for their favorable stability properties for stiff systems  \cite{ascher1997implicit,boscarino2007error,boscarino2009class,kennedy2003additive}.

\textbf{Linear Implicit System} At each stage, one needs to solve a nonlinear system for $U^{(i)}$, typically with an iterative method such as the Newton method. If $\bH$ is linear, however, \eqref{eq:ARK:ODE:compl} falls into a linear problem with respect to $U^{(i)}$:
\begin{small}
\begin{equation}
\begin{aligned}
\left( \I - \dt\,\tilde{a}_{ij}  \J \right) U^{(i)} = u_{n} + \dt \sum_{j=1}^{i-1} a_{ij} \bG^{(j)} +
\dt \sum_{j=1}^{i-1} \tilde{a}_{ij} \, \J \, U^{(j)},
\end{aligned}
\label{eqn:linearsys}
\end{equation}
\end{small}
where $\J$ is constant across the $s$ stages and all the time steps. 
Thus, the linear-nonlinear partitioning can significantly increase computational efficiency by avoiding solving a nonlinear system. In practice, this partitioning is naturally applicable to many multiphysics problems. For example, in diffusion-reaction equations, diffusion can be represented by a linear term and treated implicitly in an IMEX setting. In neural-network-based models, layers such as convolution layers or fully connected layers without activation functions can be viewed as a linear map as well.

\textbf{Efficient Mini-batching}
Consider a mini-batch $
\mathcal{B}_n = \{u_n^1,\dots,u_n^m \}
$ consisting of $m$ inputs at time $t_n$.
The batch Jacobian in \eqref{eqn:linearsys} can be represented in Kronecker product form:
\begin{small}
\begin{equation}
\begin{aligned}
\left[ \left( \I - \dt\,\tilde{a}_{ij} \J \right) \otimes \I_m \right] U|_\mathcal{B}^{(i)} & =  \mathcal{B}_{n} + \dt \sum_{j=1}^{i-1} a_{ij} \bG_\mathcal{B}^{(j)}  \\ & \quad +
\dt \sum_{j=1}^{i-1} \tilde{a}_{ij} \, (\J \otimes \I_m) \, U_\mathcal{B}^{(j)},
\end{aligned}
\label{eqn:linearsys_kron}
\end{equation}
\end{small}
where $U|_\mathcal{B}$ is the corresponding stage values for $\mathcal{B}_n$.
Since the linear problem \eqref{eqn:linearsys_kron} is equivalent to
\begin{footnotesize}
\begin{equation}
\begin{aligned}
 \left[ \I - \dt\,\tilde{a}_{ij} \J   \right] \text{reshape}( U|_\mathcal{B}^{(i)},d,m )  = \text{reshape} (\mathcal{B}_{n},d,m ) \\
 \quad + \dt \sum_{j=1}^{i-1} a_{ij} \text{reshape}(\bG_\mathcal{B}^{(j)},d,m)
 \\ + \dt \sum_{j=1}^{i-1} \tilde{a}_{ij} \, \J \, \text{reshape}( U|_\mathcal{B}^{(j)},d,m ),
\end{aligned}
\end{equation}
\end{footnotesize}
we need to solve only a reduced linear system with $m$ right-hand sides reshaped from the batched vectors.
This problem can be solved conveniently by off-the-shelf linear solvers, including direct solvers and iterative solvers.
With direct solvers, the factorization of the left-hand side $\I - \dt\,\tilde{a}_{ij} \J$ can be reused across stages and time steps because $\J$ is modified at the end of each training iteration when the NN parameters are updated by the optimizer.
If $\J$ is known a priori and not parameterized, the factorization can be reused even across mini-batches.
Iterative solvers are typically chosen because of their excellent scalability and low memory requirement. 
If $\J$ is large and sparse, using a matrix-free iterative solver may be advantageous.


\subsection{Backward Pass}

Unlike the vanilla neural ODE that employs a continuous adjoint approach, we use a discrete approach as an alternative to directly backpropagating through an ODE solver.

\begin{theorem}
The gradient of loss $\ell$ with respect to the NN parameters $p$ for SINODE using the IMEX-RK  methods in \eqref{eq:ARK:ODE:compl} can be calculated with the following discrete adjoint formula:


\begin{small}
\allowdisplaybreaks
\begin{equation}
\begin{aligned}
  \left( I - \Delta t \, \tilde{a}_{ii} \,  \bHu^{T({i})} \right) \blambda_{n+1}^{(i)}   &= \Delta t \left( b_i \bGu^{T(i)}  + \tilde{b}_i \bHu^{T(i)} \right) \blambda_{n+1} \\
     & \quad + \Delta t \,\bGu^{T(i)}  \sum_{j=i+1}^s  a_{ji} \blambda_{n+1}^{(j)}  \\
     & \quad +  \Delta t \, \bHu^{T(i)} \sum_{j=i+1}^s \tilde{a}_{ji}  \blambda_{n+1}^{(j)}, \\
     & \qquad i=s,\cdots,1,  \\
     \blambda_n &=  \blambda_{n+1} + \sum_{j=1}^s \blambda_{n+1}^{(j)}, \\
    \bmu_{n+1}^{(i)} & = \Delta t \left( b_i \bGp^{T(i)} +  \tilde{b}_i  \bHp^{T(i)} \right) \blambda_{n+1} \\
    & \quad + \Delta t \, \bGp^{T(i)}  \sum_{j=i+1}^s  a_{ji} \blambda_{n+1}^{(j)} \\
      & \quad +  \Delta t \, \bHp^{T(i)} \sum_{j=i}^s  
 \tilde{a}_{ji}  \blambda_{n+1}^{(j)}, \\
  & \qquad i=s,\cdots,1, \\
 \bmu_n  & = \bmu_{n+1} + \sum_{j=1}^s \bmu_{n+1}^{(j)},
\end{aligned}
\label{eqn:disadj_imexrk_implicit}
\end{equation}
\end{small}
with terminal condition
\begin{equation}
\blambda_N = \frac{d \ell}{d u}, \quad \bmu_N =0 .
\label{eqn:term}
\end{equation}
Here $\blambda$ and $\bmu$ correspond to the partial derivatives of the loss with respect to the initial state and the NN parameters, respectively.
\end{theorem}
At the beginning of the backward pass, we initialize $\blambda$ and $\bmu$
according to the terminal condition \eqref{eqn:term}.
Then they are propagated backward in time (the index $n$ goes from $N$ to $0$) during the backward pass according to \eqref{eqn:disadj_imexrk_implicit}.
At the end of the backward pass, we obtain the gradient 
\[
\nabla_p \ell = \bmu_0^T. 
\]
If $\tilde{a}_{ii}=0$, the first equation in \eqref{eqn:disadj_imexrk_implicit} falls back to
\begin{small}
\begin{equation}
\begin{aligned}
& \blambda_{n+1}^{(i)} =  \Delta t \left( b_i \bGu^{T(i)}  + \tilde{b}_i \bHu^{T(i)} \right) \blambda_{n+1} \\ & \quad + \Delta t \,\bGu^{T(i)} \sum_{j=i+1}^s  a_{ji} \blambda_{n+1}^{(j)} 
+ \Delta t \, \bHu^{T(i)}\sum_{j=i+1}^s 
 \tilde{a}_{ji}  \blambda_{n+1}^{(j)}.
 \end{aligned}
\end{equation}
\end{small}
For all $\tilde{a}_{ii}$ that is nonzero, we need to solve a linear system that involves the transposed Jacobian.
Nevertheless, the same strategy described in Sec. \ref{sec:fwd} can be applied to solving this transposed system. 
For efficiency, we compute the transposed Jacobian-vector products in \eqref{eqn:disadj_imexrk_implicit} with automatic differentiation tools such as AutoGrad.
Each time, the computational cost is one forward evaluation of the right-hand side $\bG$ or $\bH$ followed by a backward pass; and, more importantly,
the memory cost depends only on the NN that approximates the ODE right-hand sides, not on the iterations needed by the linear solvers or the number of time steps. 
Note that the Jacobian depends on the ODE states $u_n$ that can be stored during the forward pass.
One can checkpoint selective states and recompute the missing states from the checkpoints during the backward pass.
A checkpointing strategy \cite{Zhang2022pnode} can be used to balance memory consumption and recomputation.

The IMEX algorithms and their adjoints can be implemented in any major frameworks, such as PyTorch.
We have implemented these algorithms as off-the-shelf solvers in PETSc \cite{petsc-user-ref} to leverage the state-of-the-art linear solvers \cite{Mills2021}  and the checkpointing techniques \cite{Zhang2021iccs,Zhang2023JCS} in PETSc.
SINODE is implemented in the PNODE framework \cite{Zhang2022pnode} that integrates PyTorch and PETSc seamlessly.

\section{Experiments}\label{sec:exp}

We validate and evaluate the performance of our methods when learning stiff ODE systems.
Throughout this section we compare our methods with a variety of baseline methods including explicit and implicit methods.
The neural network architectures we use follow the best settings identified in previous works \cite{Linot2022,chamberlain2021grand}.
The only modification necessary for using our method is to split the ODE right-hand side.
Code is available at \url{https://github.com/caidao22/pnode}.

\subsection{Graph Classification with GRAND}\label{sec:grand}

GRAND is an expansive and innovative class of GNNs that draws inspiration from the discretized diffusion PDE and applies it to graphs. These architectures address prevailing challenges of current graph machine learning models such as depth, over-smoothing, bottlenecks, and graph rewiring. With a reinterpretation of the diffusion equation for graphs, the process can be depicted by an ODE system:
\begin{equation}
  \frac{d x(t)}{d t} = (\mathbf{A}(x(t))-\mathbf{I})x(t)
  \label{eqn:grand}
\end{equation}
where $\mathbf{A}(x(t)) = (a(x_i(t),x_j(t)))_{ij}$ is an attention matrix with the same structure as the adjacency matrix of the graph and $\mathbf{I}$ is an identity matrix.


We introduce an ODE block where we apply the SINODE approach and split the right-hand side of \eqref{eqn:grand} into a nonlinear part and a linear part:
\begin{equation}
  \frac{d x(t)}{d t} = (A(x(t)) - I) x(t) = \underset{\text{Nonlinear}}{\underbrace{A(x(t))x(t)}} + 
  \underset{\text{Linear}}{\underbrace{ (-I x(t)) }} . 
  \label{eqn:grand_split}
\end{equation}
The nonlinear part is handled explicitly, and the linear part is handled implicitly with IMEX-RK methods.
We use an iterative matrix-free solver in PETSc for solving the linear systems because of the simplicity of the linear part.

For assessment, we choose three benchmark datasets: Cora, CoauthorCS, and Photo.
In addition to the baseline methods mentioned, we include the implicit Adams, which is claimed to perform well for this problem in \cite{chamberlain2021grand}.
Because of the stability constraints, we have to utilize a step size of $0.005$ for explicit methods, while the implicit methods and the IMEX methods allow us to use a step size of $1$ thanks to their superior stability properties.

\begin{figure}
\centering
\includegraphics[width=0.44\textwidth, valign=t]{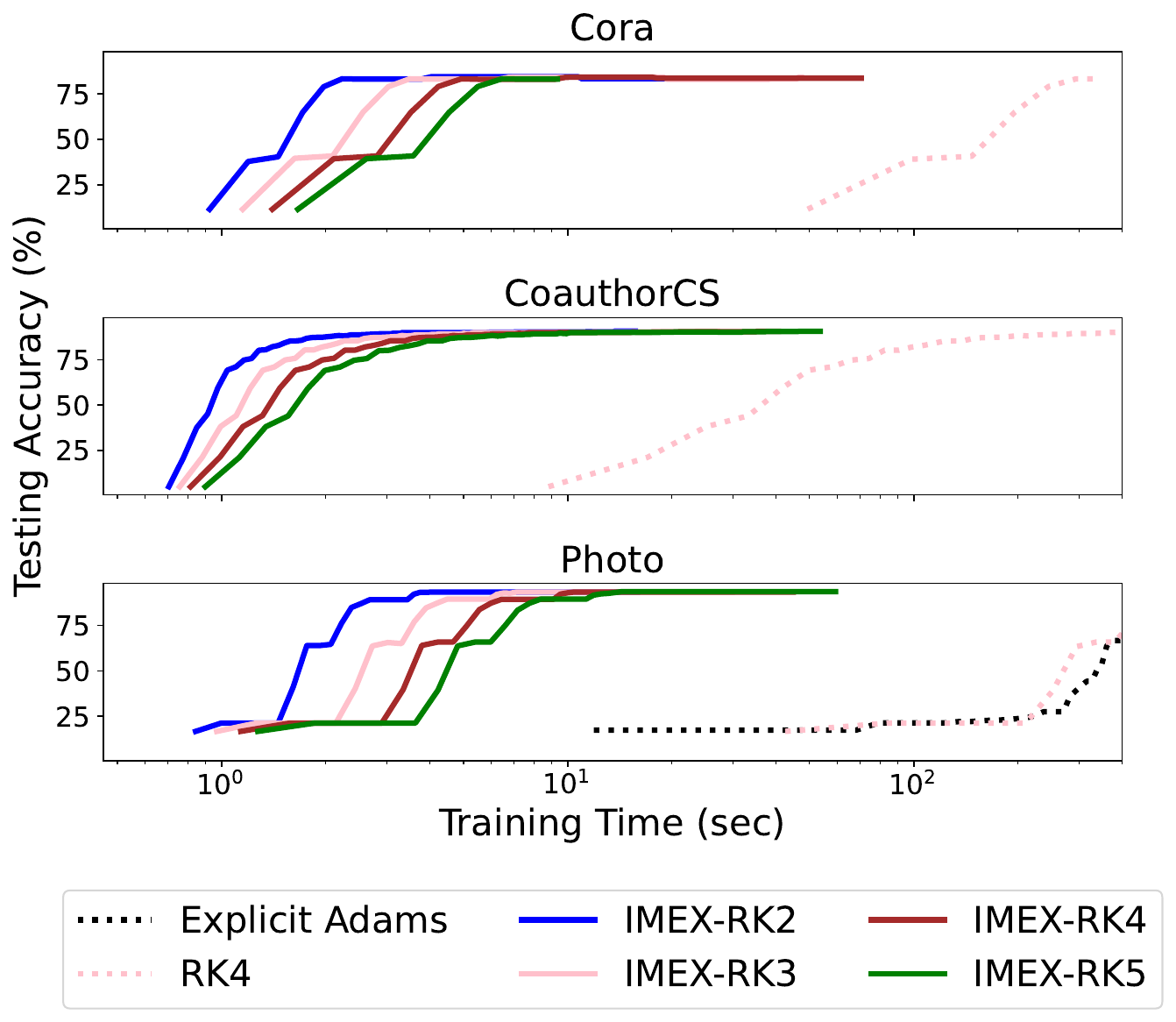}
\caption{Testing accuracy versus training time for various time integrators using \textbf{explicit} and \textbf{semi-implicit} methods.}
\label{fig:grand_clock}
\end{figure}

\begin{table}
\renewcommand{\arraystretch}{1.1} 
    \begin{center}
        \small
        \begin{tabular}{c  c  c  c} 
           \toprule
            & \multicolumn{3}{c}{NFEs} \\
            \cline{2-4}
            & Cora & CoauthorCS & Photo \\
          \midrule
RK4 & 43908 & 7512 & 8572 \\
Explicit Adams & 10,995 & 1896 & 2169 \\
Implicit Adams & 1852 & 73 & 62 \\
IMEX-RK2 & \textbf{114} & \textbf{24} & \textbf{24} \\
IMEX-RK3 & \textbf{228} & \textbf{48} & \textbf{48} \\
IMEX-RK4 & \textbf{342} & \textbf{72} & \textbf{72} \\
IMEX-RK5 & \textbf{456} & \textbf{96} & \textbf{96} \\
Crank--Nicolson & 611 & 174 & 447 \\
\bottomrule
        \end{tabular}
        \caption{NFEs per epoch for neural ODEs with explicit methods and implicit methods and SINODE with IMEX methods. 
        }
        \label{tab:grand_nfe}
    \end{center}
    \vspace{-5pt}
\end{table}
Compared with the explicit methods and the implicit methods, our approach boasts advantages under two performance evaluation metrics.
On the one hand, as depicted in Table~\ref{tab:grand_nfe}, SINODE with IMEX-RK methods requires significantly fewer numbers of right-hand side function evaluations (NFEs) than explicit methods do.
Each function evaluation corresponds strictly to one forward pass of the NN.
Thus, NFE is usually a good metric that accounts for the dominant cost for training neural ODEs while being independent of the NN complexity.
On the other hand, as illustrated in Fig.~\ref{fig:grand_clock}, SINODE with IMEX-RK methods takes much less training time than classical neural ODEs with explicit methods to achieve the same level of testing accuracy.

\begin{figure}
\centering
\includegraphics[width=0.44\textwidth, valign=t]{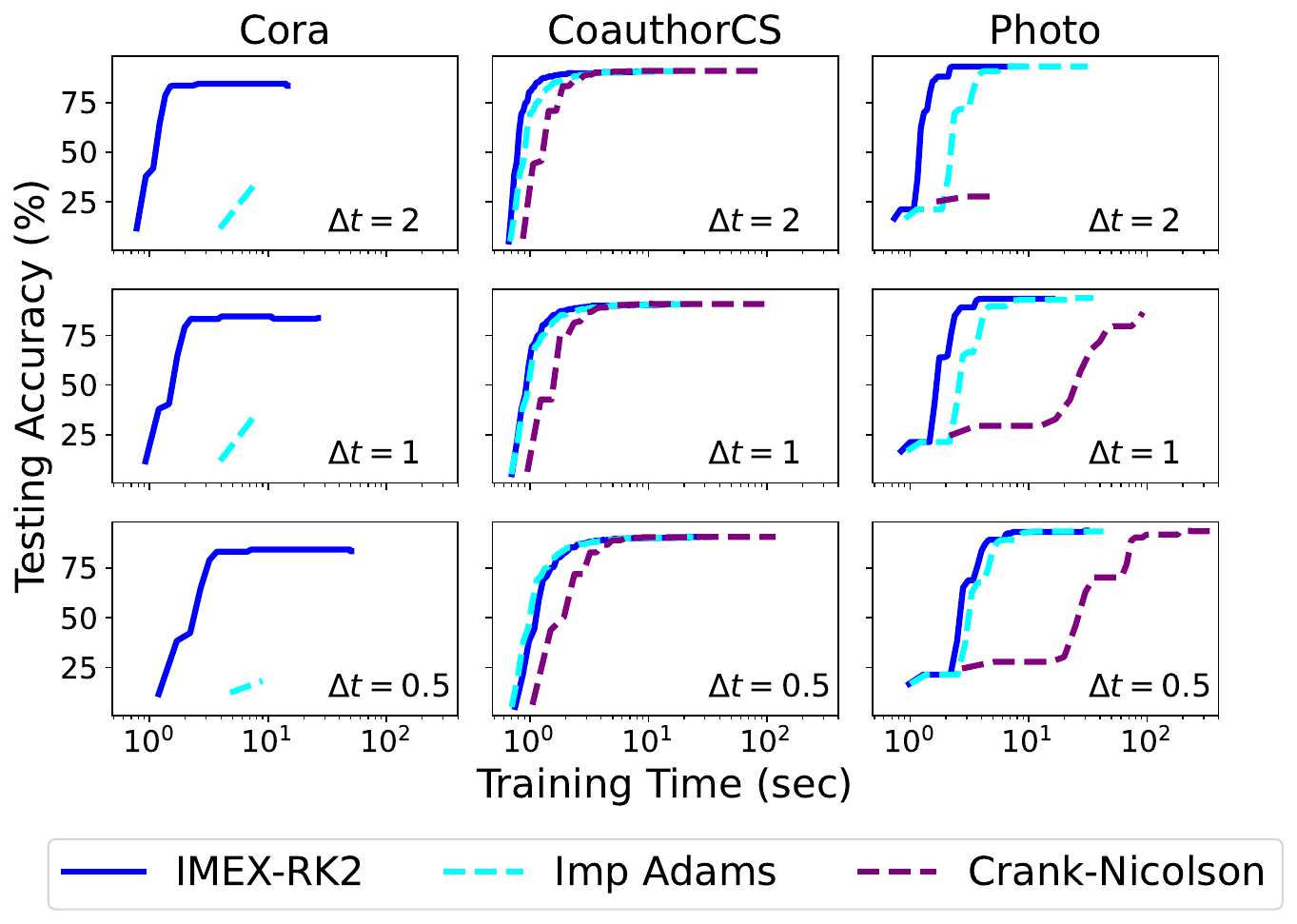}
\caption{Testing accuracy vs. training time for various step sizes for \textbf{implicit} and \textbf{semi-implicit} methods.
For Cora, Crank--Nicolson failed in the beginning of training, and implicit Adams failed in the middle of training.}
\label{fig:grand_clock_implicit}
\end{figure}

We have also performed a comparison between IMEX-RK2 and fully implicit methods for varying step sizes.  Fig.~\ref{fig:grand_clock_implicit} shows that IMEX-RK2 outperforms the implicit Adam and Crank--Nicolson significantly for Cora and Photo.
Both implicit Adam and Crank--Nicolson cannot complete the training for Cora because the loss blows up.

\subsection{Learning Dynamics for the Kuramoto–Sivashinsky (KS) Equation} \label{sec:kse}
The KS equation in one-dimensional space can be written as
\begin{equation}
\frac{\partial u}{\partial t}  = - u \frac{\partial u} {\partial x} - \frac{\partial^2 u}{\partial x^2} - \frac{\partial^4 u}{\partial x^4}, \quad x \in [0, 22].
\label{eq:KS}
\end{equation}
It is a well-known nonlinear \emph{chaotic} PDE for modeling complex spatiotemporal dynamics. 

We train neural ODEs by approximating the nonlinear term $u \frac{\partial u} {\partial x}$ with a fully connected multilayer perceptron (MLP) and approximating the remaining linear terms on the right-hand side of \eqref{eq:KS} with a fixed convolutional NN (CNN) layer
. 
We use circular padding and a CNN filter of size $5$ to mimic centered finite-difference operators for the antidiffusion and hyperdiffusion terms with periodic boundary conditions.
The parameters of the filter are fixed to be $[-1.0/\Delta x^4, 4.0/\Delta x^4-1.0/\Delta x^2, -6.0/\Delta x^4+2.0/ \Delta x^2, 4.0/ \Delta x^4-1.0/ \Delta x^2, -1.0/ \Delta x^4]$, where the grid spacing $\Delta x = L/N$.
A direct solver with reused LU factorizations is used to solve the linear systems associated with the CNN layer.

\begin{figure}[ht]
  \centering
  \includegraphics[width=0.49\linewidth]{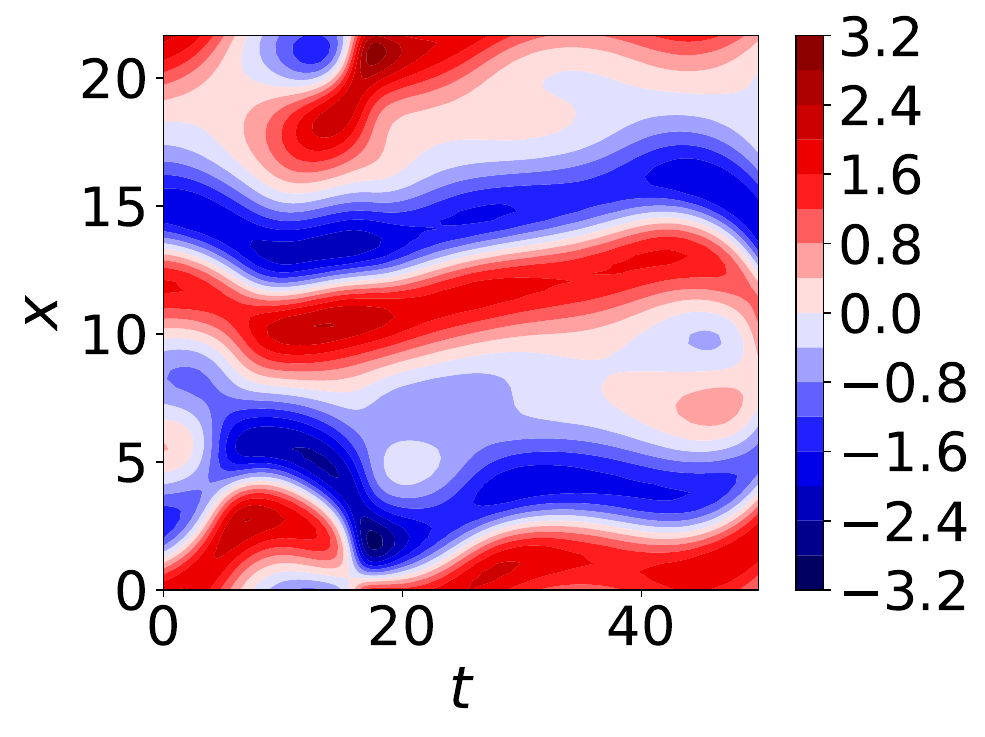}
  \hfill
  \includegraphics[width=0.49\linewidth]{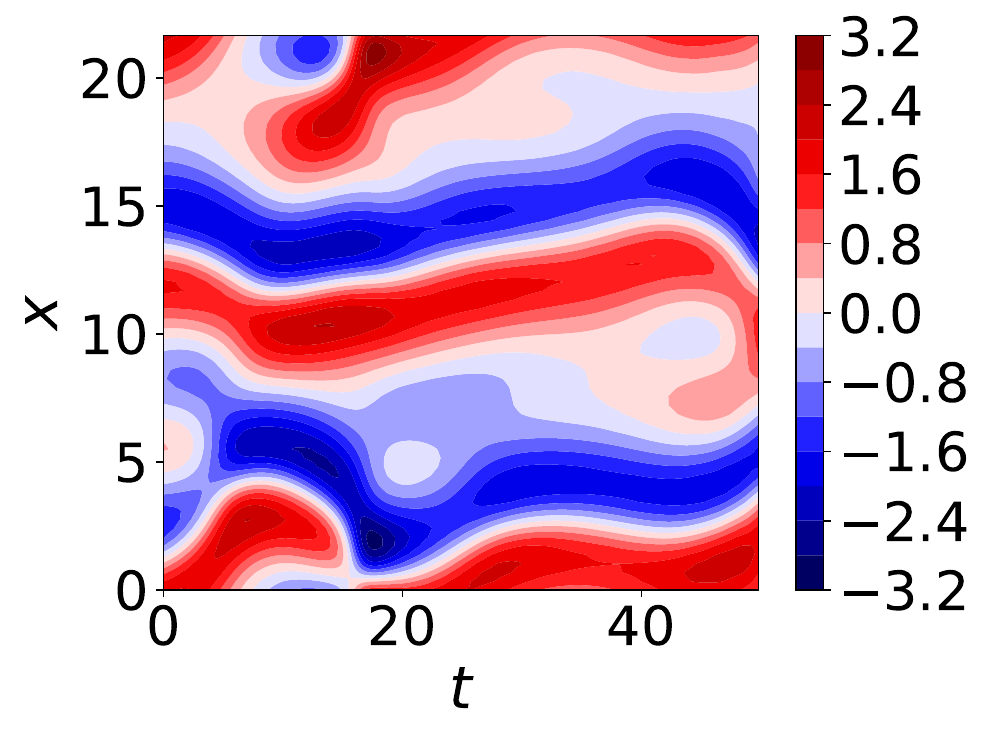}
  \caption{Ground truth (\textbf{Left}) vs. prediction (\textbf{Right})  using IMEX-RK3 with $\Delta t = 0.2$.}
  \label{fig:kse_gt}
\end{figure}
%
When applying SINODE, we treat the MLP part explicitly and the CNN part implicitly.
A time step size $0.2$ is used for the IMEX methods and the fully implicit method.
Because of the stability constraints, the explicit methods do not work well until we decrease the step size to $0.001$.
Fig.~\ref{fig:kse_gt} compares a true trajectory to predictions from SINODE.
We can see that the predicted trajectories stay on the attractor for a long time and match the ground truth well.

\begin{figure*}
  \centering
  \includegraphics[width=0.6\linewidth]{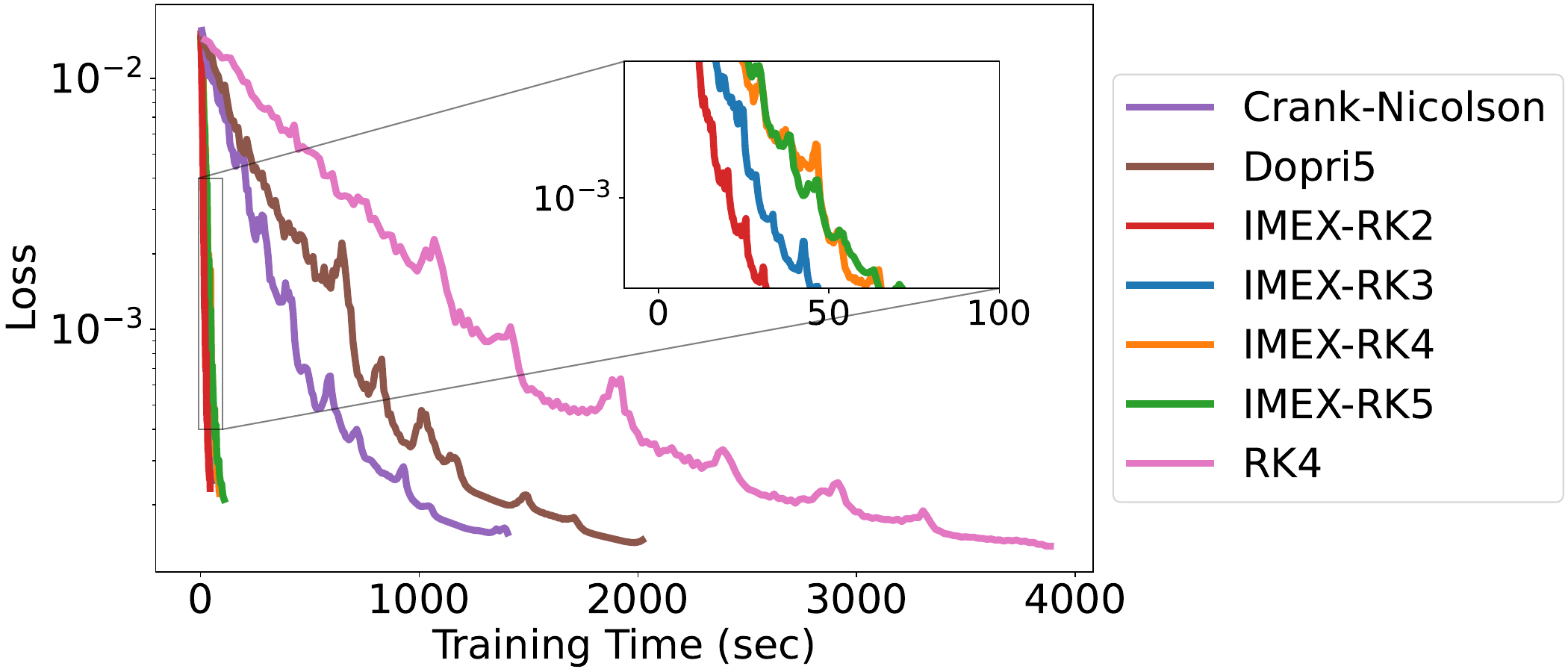}
  \hfill
  \includegraphics[width=0.38\linewidth]{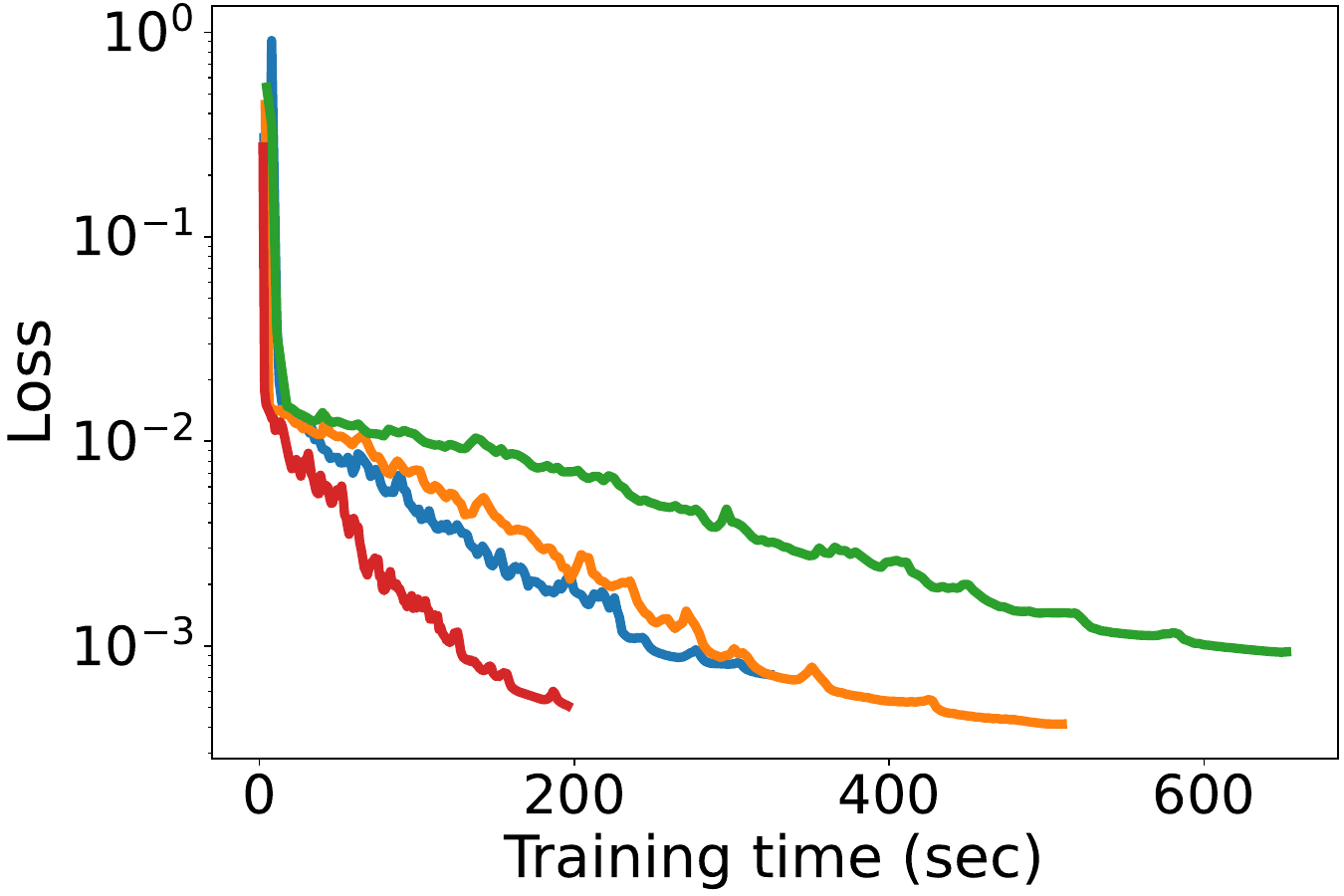}
  \caption{Training loss vs. training time for KS. \textbf{Left}: Grid size 64. \textbf{Right}: Grid size 512. Only SINODE with IMEX methods achieve feasible training completion for the case with grid size 512.}
  \label{fig:kse_time64}
\end{figure*}

In Fig.~\ref{fig:kse_time64} we show the training loss versus the training time for all the methods in order to evaluate their efficiency.
As expected, SINODE using IMEX methods significantly outperforms the other methods.
For example, SINODE with IMEX-RK2 is $\sim$$47$ times faster than the classical neural ODE with Dopri5 and $\sim$$33$ times faster than the neural ODE with Crank--Nicolson to decrease the training loss to $10^{-3}$.
Table~\ref{tab:pde_nfe} compares the average NFEs per epoch for different solvers.
It shows that SINODE using IMEX-RK methods can reduce the NFEs remarkably.
The superior performance of SINODE is mainly attributed to superior stability properties and the computational advantages of semi-implicit methods.
Note that the improvement in training efficiency becomes more significant as the number of time steps increases because of the increased opportunity to reuse the LU factorization for solving the linear systems.
In our experiments we use only one time step during the training, which reflects the worst-case performance of SINODE.

Training with neural ODEs becomes more difficult when the grid resolution increases.
After we change the grid size from $64$ to $512$, the performance of neural ODEs with explicit methods or fully implicit methods degrades dramatically while SINODE maintains high efficiency.
We can still use the time step size $0.2$ for the training of SINODE, but the explicit methods require a step size on the order of $10^{-7}$ because of the severe stability constraint from the hyperdiffusion term.
We are not able to make the fully implicit methods work with the step size $0.2$ because the nonlinear solver diverges frequently during the training process.
This is not uncommon for classical PDE solvers since the condition number of the system increases as the grid resolution increases.
Decreasing the step size by $10$ times stabilizes the nonlinear solvers.
However, this leads to a per-epoch training time of more than one hour whereas SINODE typically takes less than one second per epoch.

An interesting observation is that the lower-order IMEX-RK methods lead to better training efficiency than high-order IMEX-RK methods, despite their slightly slower convergence speed.
This is because higher-order IMEX-RK methods require more stages, leading to more linear solves at each time step.
For the coarse-grid case ($64$), the second-order IMEX-RK performs $\sim$$2.5$ times better than the fifth-order IMEX-RK.
For the fine-grid case ($512$),  its gain is larger than a factor of $5$.
We expect that the gain to increase as the grid resolution further increases.

\subsection{Learning Dynamics for the Viscous Burgers Equation} \label{sec:burgers}

We consider the one-dimensional viscous Burgers equation
\begin{equation}
    \frac{\partial u}{\partial t}  = - u \frac{\partial u} {\partial x} + \nu \frac{\partial^2 u}{\partial x^2}, \quad x \in [0, 1],
    \label{eqn:VB}
\end{equation}
with viscosity $\nu = 8\cdot10^{-4}$, which makes the solution known as Burger turbulence.

The architecture of the neural ODEs approximates the nonlinear term $- u \frac{\partial u} {\partial x}$ with a fully connected nonlinear NN, and the linear term $\nu \frac{\partial^2 u}{\partial x^2}$ with a fixed linear layer derived from a finite difference method.
Again we choose the direct solver to benefit from the reuse of LU factorizations.
In the comparison of different methods, we choose a time step size of $0.05$ as a conservative choice for the four IMEX-RK methods and the fully implicit method. Explicit methods cause the solution to explode until the time step size decreases to as small as $0.001$ when the grid size is $512$ and to $0.0005$ when the grid size is $1024$.

\begin{figure}[H]
\centering
  \includegraphics[width=\linewidth]{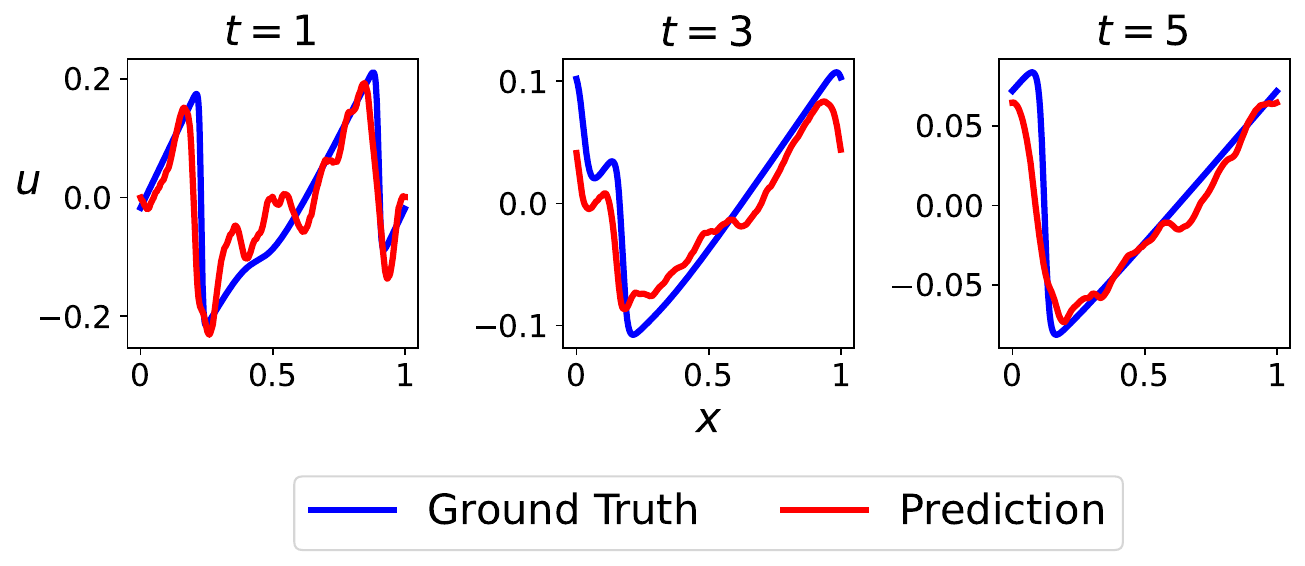}
\caption{Snapshots of VBE trajectories 
for grid size $512$ using IMEX-RK3.
}
\label{fig:burgers_snapshot_512_IMEX}
\vspace{-5pt}
\end{figure}

\begin{figure*}
\centering
\includegraphics[width=0.57\linewidth]{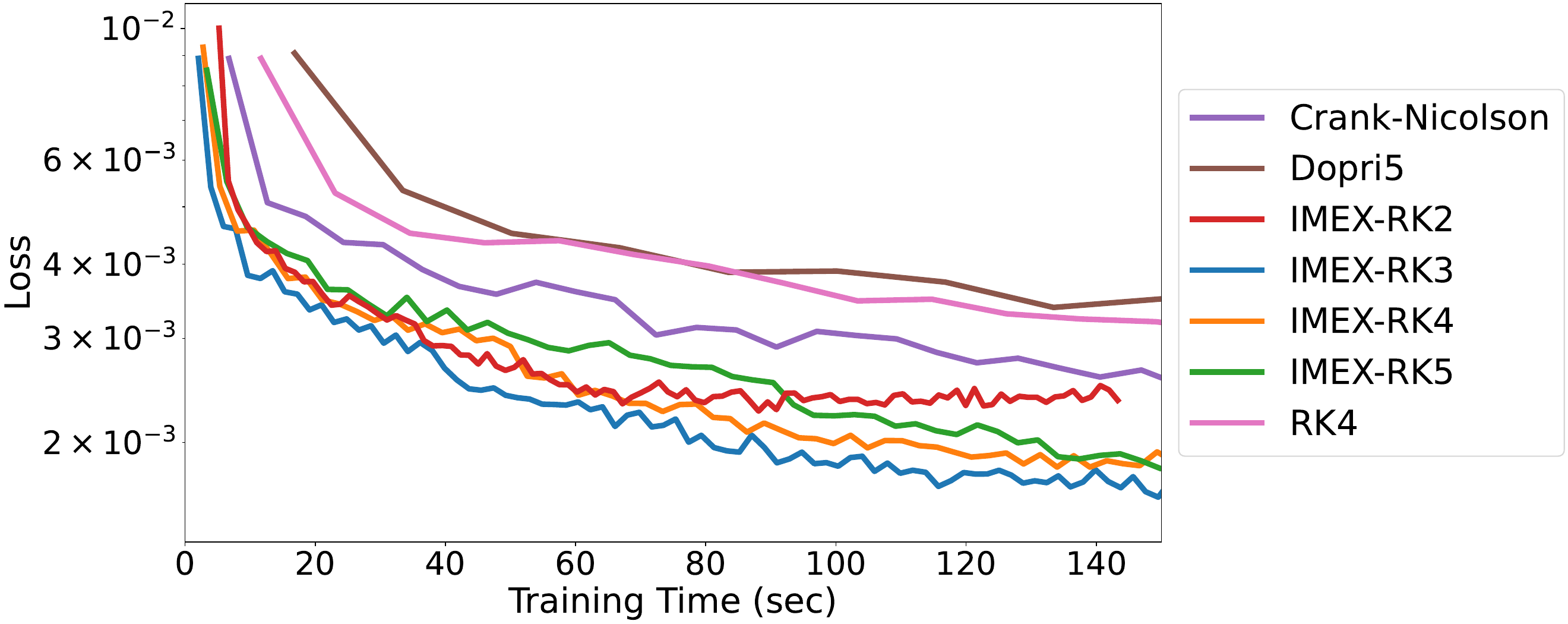}
\hfill
\includegraphics[width=0.42\linewidth]{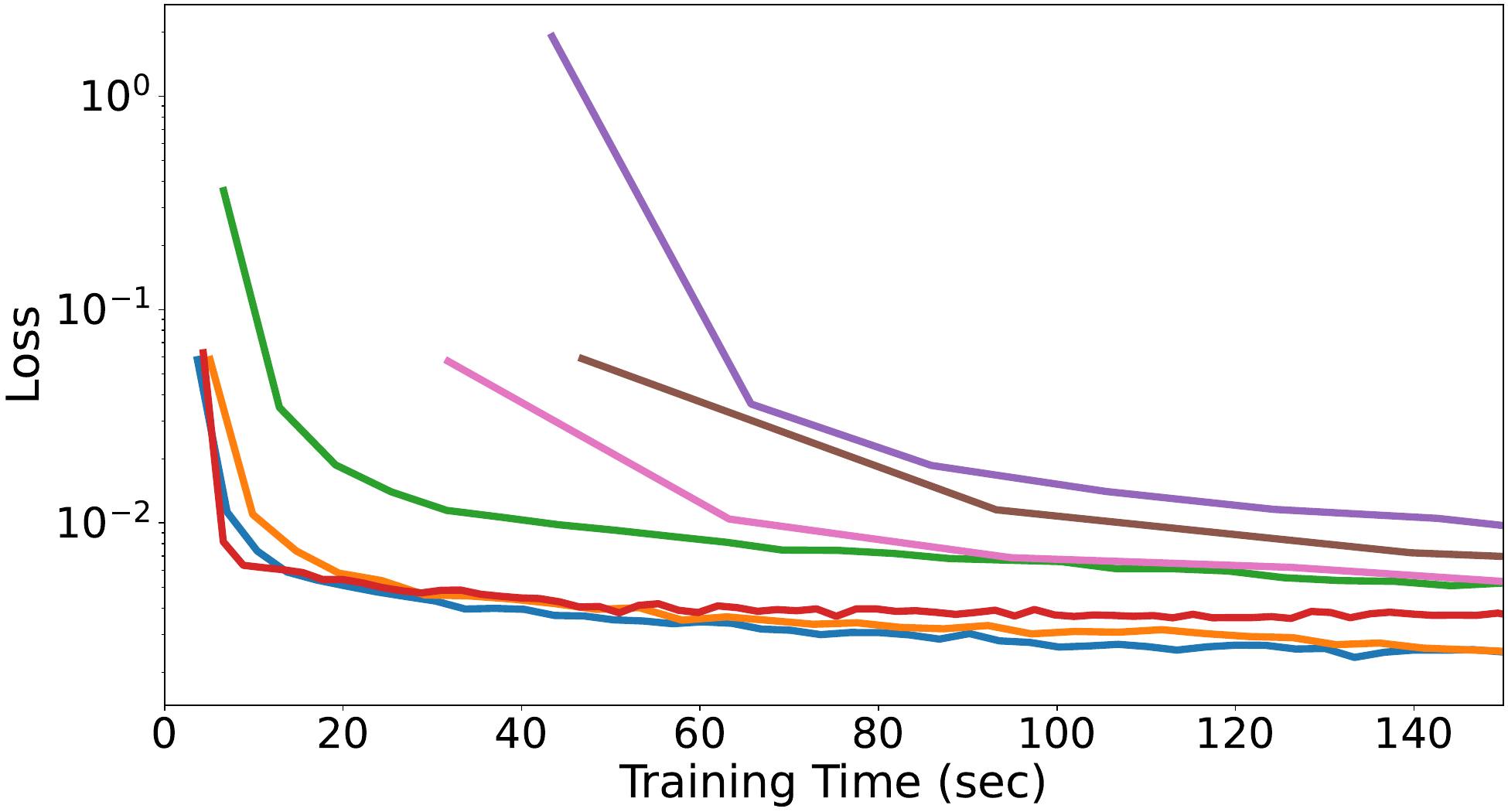}
\caption{Training loss vs. training time for VBE. \textbf{Left}: Grid size 512. \textbf{Right}: Grid size 1024.
The explicit Adams does not converge due to instability and is thus not shown.}
\label{fig:burgers_loss_512}
\end{figure*}

Fig.~\ref{fig:burgers_snapshot_512_IMEX} shows that the predictions made by SINODE are in good agreement with the true solutions for an initial condition from the testing dataset.
\begin{table}
\renewcommand{\arraystretch}{1.2} 
    \begin{center}
        \small
        \begin{tabular}{ c c c c c}
        \toprule
            & \multicolumn{4}{c}{NFEs} \\
            \cline{2-5}
            & \multicolumn{2}{c}{KSE}  & \multicolumn{2}{c}{Burgers}   \\
            & grid 64 & grid 512 & grid 512 & grid 1024 \\ 
            \midrule
            Dopri5 & 12619 & $-$ & 22819 & 45619 \\
            RK4 & 24000 & $-$ & 15200 & 30400 \\
            IMEX-RK2 & \textbf{60} &  \textbf{60} & \textbf{152} & \textbf{152} \\
            IMEX-RK3 & \textbf{120} & \textbf{120} & \textbf{304} & \textbf{304} \\
            IMEX-RK4 & \textbf{180} & \textbf{180}& \textbf{456} &\textbf{456} \\
            IMEX-RK5 & \textbf{240} & \textbf{240} & \textbf{608} &\textbf{608} \\
            Crank--Nicolson & 3195 &$-$ & 13072 & 11883 \\
\bottomrule
        \end{tabular}
        \caption{NFEs for neural ODEs with explicit methods and implicit methods and SINODE with IMEX methods. The numbers are rounded to the closest integer.}
        \label{tab:pde_nfe}
    \end{center}
    \vspace{-10pt}
\end{table}
Fig.~\ref{fig:burgers_loss_512} show that SINODE significantly outperforms the classical neural ODE for the two different grid settings. IMEX-RK3 is consistently the best-performing method.
For the case with grid size $512$,
IMEX-RK3 is $\sim 6X$ faster than RK4 in terms of training time per epoch.
The speedups increase to $\sim 10X$ for the case with grid size $1024$.
The speedups over Crank--Nicolson are $\sim 3.4X$ and $\sim 2X$ for the two cases.
The gain in time-to-accuracy
appears to be larger than $2X$ because the training converges slower with Crank--Nicolson than with IMEX-RK3.
Table~\ref{tab:pde_nfe} lists NFEs for different ODE solvers.

\section{Related Work}

\textbf{Implicit Networks}
Whereas explicit networks consist of a sequence of layers, implicit networks define the output as a solution to a nonlinear system 
For example, the well-known neural ODE model \cite{Chen2018} solves an ODE problem in the forward pass, typically with explicit integration methods such as RK methods. The DEQ model \cite{Winston2020, Bai2019} finds an equilibrium solution to a nonlinear system with a fixed-point iteration method
in the forward pass.
These models require solving a nonlinear system by embedding an iterative solver in the architecture.
Therefore, common numerical issues associated with the solver, such as \emph{convergence} and \emph{stability}, may be more frequently encountered during the training process.
For example, explicit methods can be restricted to \emph{infinitely small} step sizes when neural ODEs are applied to stiff problems.
Fully implicit methods \cite{Zhang2022pnode} are \emph{absolutely stable}, but the nonlinear solvers 
may fail to converge for ill-conditioned systems.
In addition, solving the nonlinear system poses a computational bottleneck for training implicit networks.
These limitations have motivated a variety of acceleration techniques such as IMEXnet \cite{haber19a}, equation scaling \cite{kim2021stiff}, JFB \cite{Fung2022} and proximal implicit neural ODE \cite{baker2022proximal}, as well as this work.

\textbf{Gradient Calculation for Implicit Networks}
Using backpropagation directly in the backward pass
is often intractable for training implicit networks because they may require a large number of iterations and the memory to store the computation graphs increases with this number.
In order to address this issue, adjoint methods \cite{Chen2018,Winston2020,zhuang2020} have been widely used to compute the gradient.
When applied to ODEs, these methods typically require storing selective states for the Jacobian evaluation during a forward pass, known as checkpointing, and restoring missing states during a backward pass via recomputation \cite{Chen2018,zhuang2020} or interpolation \cite{Daulbaev2020}. 

\textbf{Reverse-accurate Neural ODEs}
Although adjoint methods provide a memory-efficient alternative to direct backpropagation, 
they do not always generate an accurate gradient.
Various researchers have \cite{onken2020discretizeoptimize, Zhang2022pnode} pointed out that the vanilla neural ODE \cite{Chen2018} adopts a continuous adjoint approach that is not reverse accurate.
In particular, the gradient is not consistent with the forward pass, even if the same time integrator and the same step size are used for solving the continuous adjoint equation \eqref{eqn:adjoint_ode} as shown in \cite{Zhang2022pnode}.
Reverse-accurate variants of neural ODEs have been developed in \cite{ANODE, ANODEV2, zhuang2020, zhuang2021mali,Matsubara2023}.
These variants rely on 
backpropagation, 
checkpointing and recomputation to alleviate the memory cost. 
\citet{Zhang2022pnode} proposed a \emph{discrete adjoint} approach that achieves reverse accuracy and can leverage optimal checkpointing schedules \cite{Zhang2021iccs,Zhang2023JCS} to balance the trade-off between recomputation and memory usage.

\section{Conclusion}

We propose a semi-implicit approach, named SINODE, for learning stiff neural ODEs. 
Our approach leverages a nonlinear-linear partitioning of the ODE right-hand side and a differentiable IMEX solver that provides reverse-accurate gradient computation.
The excellent stability property of IMEX methods enables large time steps to be used for integrating the ODEs, which is especially critical for stiff problems where other methods are inefficient. 
In addition, our approach allows us to use off-the-shelf linear solvers in mini-batch training, opening the opportunity of adapting well-established techniques from the other fields. 
We demonstrate the advantages of SINODE when applied to graph learning tasks and time series modeling of two challenging dynamical systems.
We show that SINODE typically allows us to use a time step that is two orders of magnitude larger than that allowed by explicit RK methods, leading to significant speedup over classic neural ODEs.
We also show that SINODE succeeds for very stiff cases where existing methods fail due to instability.

\textbf{Limitations} SINODE should not be used for nonstiff problems where classic neural ODEs with explicit methods could be more efficient.
When direct linear solvers are used, varying step sizes may degrade the performance of SINODE because the Jacobian cannot be reused across time steps. However, the performance is not affected much when matrix-free iterative linear solvers are used.

\newpage

\section*{Acknowledgment}
This work was supported in part by the U.S. Department of
Energy, Office of Science, Office of Advanced Scientific Computing Research,
Scientific Discovery through Advanced Computing (SciDAC) program through the
FASTMath Institute and DOE-FOA-2493 "Data-intensive scientific machine learning", under contract DE-AC02-06CH11357 at Argonne National
Laboratory. We gratefully acknowledge the computing resources provided and operated by the Joint Laboratory for System Evaluation (JLSE) at Argonne National Laboratory.

\bibliography{reference.bib}

\onecolumn
\appendix

\section{Proof of Theorem 5.1}
\begin{proof}
The discrete adjoint of IMEX-RK can be derived by using Lagrangian.
To avoid a lengthy presentation, we describe the steps to derive the adjoint for an s-stage explicit RK (ERK) method represented by 
\begin{equation}
  \label{eqn:adjsensi}
  \begin{aligned}
  \bU_i & = \bu_n + \sum_{j=1}^{i-1} h_n \, a_{ij} \, F(\bU_j), \quad i=1,\cdots,s, \\
  \bu_{n+1} & = \bu_n + \sum_{i=1}^s h_n \, b_i \, F(\bU_i),
  \end{aligned}
\end{equation}
where $\bu \in \mathcal{R}^m$ and $F: \mathcal{R}^m \rightarrow \mathcal{R}^m$.

IMEX-RK can be viewed as an additive combination of two RK methods.
So the extension from ERK to IMEX-RK can be done straightforwardly by replacing
\[
  \sum_{j=1}^{i-1} \, a_{ij} \, F(\bU_j)
\]
with
\[
  \sum_{j=1}^{i-1} a_{ij} \bG^{(j)} + \sum_{j=1}^{i} \tilde{a}_{ij} \bH^{(j)} 
\]
, replacing
\[
  \sum_{j=1}^{i-1} \, b_{j} \, F(\bU_j)
\]
with
\[
  \sum_{j=1}^{i-1} b_{j} \bG^{(j)} + \sum_{j=1}^{i} \tilde{b}_{j} \bH^{(j)}, 
\]
and expanding their derivatives correspondingly.
While the extension is tedious and leads to a lengthy description, the principle stays the same for both methods.
Therefore we use ERK for illustration below.

Define the Lagrangian
\begin{equation}
\begin{aligned}
\mathcal{L}(\I) = \psi(\bu_N)  - \lambda_0 ^T \left( \bu_0 - \I \right) - \sum_{n=0}^{N-1} \lambda_{n+1} ^T \left( \bu_{n+1} -  \bu_n + \sum_{i=1}^s h_n \, b_i \, F(\bU_{n,i}) \right) \\
- \sum_{n=0}^{N-1} \sum_{i=1}^{s} \lambda_{n+1,i}^T \left( \bU_{n,i} -  \bu_n - \sum_{j=1}^{i-1} h_n \, a_{ij} \, F(\bU_{n,j}) \right),
\end{aligned}
\label{eqn:lag}
\end{equation}
where $\lambda \in \mathcal{R}^m$ is a column vector so that $\lambda^T$ is a row vector.

Take the derivative to $\I$
\begin{equation}
\begin{aligned}
\frac{d \mathcal{L}(\I)}{d \I} = \frac{d \psi(\bu_N)}{d \bu_N} \frac{\partial \bu_N}{\partial \I}  - \lambda_0 ^T \left( \frac{\partial \bu_0}{\partial \I} - I \right) - \sum_{n=0}^{N-1} \lambda_{n+1} ^T \left( \frac{\partial \bu_{n+1}}{\partial \I} -  \frac{\partial \bu_n}{\partial \I} + \sum_{i=1}^s h_n \, b_i \, \frac{\partial F(\bU_{n,i})}{\partial \bU_{n,i}}  \frac{\partial \bU_{n,i}}{\partial \I} \right) \\
- \sum_{n=0}^{N-1} \sum_{i=1}^{s} \lambda_{n+1,i}^T \left( \frac{\partial \bU_{n,i}}{\partial \I} -  \frac{\partial \bu_n}{\partial \I} - \sum_{j=1}^{i-1} h_n \, a_{ij} \, \frac{\partial F(\bU_{n,j})}{\partial \bU_{n,j}} \frac{\partial \bU_{n,j}}{\partial \I} \right).
\end{aligned}
\label{eqn:dlag}
\end{equation}
Expand the third and fourth terms on the right-hand side, and reorganize
\begin{equation}
\begin{aligned}
\frac{d \mathcal{L}(\I)}{d \I} = \left( \frac{d \psi(\bu_N)}{d \bu_N} - \lambda_N^T \right) \frac{\partial \bu_N}{\partial \I}  + \lambda_0 ^T  \\
- \sum_{n=0}^{N-1} \lambda_{n} ^T  \frac{\partial \bu_{n}}{\partial \I} + \sum_{n=0}^{N-1}  \lambda_{n+1} ^T  \frac{\partial \bu_{n}}{\partial \I} + \sum_{n=0}^{N-1} \sum_{i=1}^{s} \lambda_{n+1,i}^T \frac{\partial \bu_n}{\partial \I} \\
- \sum_{n=0}^{N-1}  \lambda_{n+1} ^T  \left( \sum_{i=1}^s h_n \, b_i \, \frac{\partial F(\bU_{n,i})}{\partial \bU_{n,i}}  \frac{\partial \bU_{n,i}}{\partial \I} \right) \\
 - \sum_{n=0}^{N-1} \sum_{i=1}^{s} \lambda_{n+1,i}^T \left( \frac{\partial \bU_{n,i}}{\partial \I} - \sum_{j=1}^{s} h_n \, a_{ij} \, \frac{\partial F(\bU_{n,j})}{\partial \bU_{n,j}} \frac{\partial \bU_{n,j}}{\partial \I} \right).
\end{aligned}
\label{eqn:dlag_reorg}
\end{equation}
We replace $\sum_{j=1}^{i-1}$ with $\sum_{j=1}^{s}$ for the convenience of changing the index later.
The two expressions are equivalent because $a_{ij} = 0 \; \forall i\leq j$. 

If we define the adjoint equations
\begin{equation} 
\label{eqn:disadj_erk2}
\begin{aligned}
  \lambda_N^T &= \frac{d \psi(\bu_N)}{d \bu_N} \\
  \blambda_n^T & = \blambda_{n+1}^T + \sum_{j=1}^s \blambda_{n+1,j}^T, \\
  \lambda_{n+1} ^T  b_i \, \frac{\partial F(\bU_{n,i})}{\partial \bU_{n,i}} & =  \lambda_{n+1,i}^T  \left( I  - \sum_{j=1}^{s} h_n \, a_{ji} \, \frac{\partial F(\bU_{n,i})}{\partial \bU_{n,i}} \right), i = s,\cdots,1
\end{aligned}
\end{equation}
we have
\begin{equation}
\frac{d \mathcal{L}(\I)}{d \I} =  \lambda_0 ^T. 
\end{equation}

We rewrite the adjoint equations \eqref{eqn:disadj_erk2} using column vectors for the ease of implementation:
\begin{equation}
\label{eqn:disadj_erk_simp}
\begin{aligned}
  \blambda_{n+1,i} & =  h_n \fu^T(\bU_i)  \left( b_i \blambda_{n+1} + \sum_{j=i+1}^s a_{ji} \, \blambda_{n+1,j} \right) , \quad i=s,\cdots,1,\\
  \blambda_n & = \blambda_{n+1} + \sum_{j=1}^s \blambda_{s,j}. \\
\end{aligned}
\end{equation}
Here we take into account that $a_{ji} = 0 \, \forall j \leq i$.

\begin{equation}
\label{eqn:disadj_erk}
\begin{aligned}
  \blambda_{s,i} & =  h_n \fu^T(\bU_i)  \left( b_i \blambda_{n+1} + \sum_{j=i+1}^s a_{ji} \, \blambda_{s,j} \right) + h_n \, b_i\,r_{\bu}^T(\bU_i)), \quad i=s,\cdots,1,\\
  \bmu_{s,i} & = h_n \fp^T(\bU_i) \left( b_i \blambda_{n+1} + \sum_{j=i+1}^s a_{ji} \, \blambda_{s,j} \right) + h_n \, b_i\,r_{\bp}^T(\bU_i), \quad i=s,\cdots,1,\\
  \blambda_n & = \blambda_{n+1} + \sum_{j=1}^s \blambda_{s,j}, \\
  \bmu_N & = \bmu_{n+1} + \sum_{j=1}^s \bmu_{s,j}  
\end{aligned}
\end{equation}
\end{proof}

\begin{remark}
To reduce the matrix-vector multiplications, we implement the reorganized version of \eqref{eqn:disadj_imexrk_implicit} as shown below. 
\begin{equation}
\label{eqn:disadj_imexrk_implicit_reorg}
\begin{aligned}
  \left( I - \Delta t \, \tilde{a}_{ii} \,  \bHu^{T({i})} \right) \blambda_{n+1}^{(i)} & = 
  \Delta t \bGu^{T(i)}  \left( b_i \blambda_{n+1} + \sum_{j=i+1}^s  a_{ji} \blambda_{n+1}^{(j)} \right) \\
  & + \Delta t \bHu^{T(i)} \left( \tilde{b}_i \blambda_{n+1} + \sum_{j=i+1}^s 
 \tilde{a}_{ji}  \blambda_{n+1}^{(j)} \right) , \quad i=s,\cdots,1,\\
  \blambda_n & = \blambda_{n+1} + \sum_{j=1}^s \blambda_{n+1}^{(j)},\\
    \bmu_{n+1}^{(i)} & = \Delta t \bGp^{T(i)}  \left( b_i  \blambda_{n+1} +  \sum_{j=i+1}^s  a_{ji} \blambda_{n+1}^{(j)}  \right) \\
  & + \Delta t \, \bHp^{T(i)} \left( \tilde{b}_i \blambda_{n+1} +   \sum_{j=i}^s 
 \tilde{a}_{ji}  \blambda_{n+1}^{(j)}  \right) , \quad i=s,\cdots,1,\\
 \bmu_n & = \bmu_{n+1} + \sum_{j=1}^s \bmu_{n+1}^{(j)}. \\
\end{aligned}
\end{equation}
\end{remark}

\newpage
\section{Coefficients of the IMEX-RK methods}
\renewcommand{\arraystretch}{1.4}
IMEX-RK methods can be represented by a pair of Butcher tableaux. 
\begin{equation}
\begin{array}{c|c c c c}
c_1 & a_{11} & a_{12} & \dots & a_{1s} \\
c_2 & a_{21} & a_{22} & \dots & a_{2s} \\
\vdots & \vdots & \vdots & \ddots & \vdots \\
c_s & a_{s1} & a_{s2} & \dots & a_{ss} \\
\hline
    & b_{1} & b_{2} & \dots & b_{s} \\
\end{array}
\qquad
\begin{array}{c|c c c c}
\tilde{c}_1 & \tilde{a}_{11} & \tilde{a}_{12} & \dots & \tilde{a}_{1s} \\
\tilde{c}_2 & \tilde{a}_{21} & \tilde{a}_{22} & \dots & \tilde{a}_{2s} \\
\vdots & \vdots & \vdots & \ddots & \vdots \\
\tilde{c}_s & \tilde{a}_{s1} & \tilde{a}_{s2} & \dots & \tilde{a}_{ss} \\
\hline
    & \tilde{b}_{1} & \tilde{b}_{2} & \dots & \tilde{b}_{s} \\
\end{array}
\end{equation}
The coefficients of all the IMEX-RK methods used in this paper are given below (if $c$ is omitted, it is the sum of the corresponding row in $A$).

\textbf{IMEX-RK2} \cite{pareschi2005implicit}
\begin{equation}
\begin{array}{c|c c}
0 & 0 & 0  \\
1 & 1 & 0  \\
\hline
    & 0.5 & 0.5 \\
\end{array}
\qquad
\begin{array}{c|c c}
1-1/\sqrt{2} & 1-1/\sqrt{2} & 0  \\
1/\sqrt{2} & 2/\sqrt{2} -1 & 1-1/\sqrt{2}  \\
\hline
    & 0.5 & 0.5 \\
\end{array}
\end{equation}

\textbf{IMEX-RK3} \cite{KENNEDY2003}
\begin{equation}
\begin{aligned}
&
\begin{array}{c|c c c c}
  & 0 & 0 & 0 & 0 \\
  & \frac{1767732205903}{2027836641118} & 0 & 0 & 0 \\
  & \frac{5535828885825}{10492691773637} & \frac{788022342437}{10882634858940} &  0 & 0 \\
  & \frac{6485989280629}{16251701735622} & -\frac{4246266847089 }{9704473918619} & \frac{10755448449292}{10357097424841} & 0 \\
\hline
& \frac{1471266399579}{7840856788654} &  -\frac{4482444167858} {7529755066697} & \frac{11266239266428}{11593286722821} & \frac{1767732205903} {4055673282236} \\
\end{array}
\\
&
\begin{array}{c|c c c c}
    & 0 & 0 & 0 &  0  \\
    & \frac{1767732205903}{4055673282236} & \frac{1767732205903}{4055673282236} &  0 &   0   \\
    & \frac{2746238789719}{10658868560708} & -\frac{640167445237}{ 6845629431997} &  \frac{1767732205903}{4055673282236} &   0 \\
    & \frac{1471266399579}{7840856788654} &  -\frac{4482444167858} {7529755066697} & \frac{11266239266428}{11593286722821} & \frac{1767732205903} {4055673282236} \\
\hline
& \frac{1471266399579}{7840856788654} &  -\frac{4482444167858} {7529755066697} & \frac{11266239266428}{11593286722821} & \frac{1767732205903} {4055673282236} \\
\end{array}
\end{aligned}
\end{equation}

\textbf{IMEX-RK4} \cite{KENNEDY2003}
\begin{equation}
\begin{aligned}
&
\begin{array}{c |c c c c c c}
 & 0 & 0 & 0 & 0 & 0 &  0 \\
 &
  \frac{1}{2} & 0 & 0 & 0 & 0 & 0 \\
 & \frac{13861}{62500} & \frac{6889} {62500} & 0 & 0 & 0 & 0 \\
 & -\frac{116923316275}{2393684061468} & -\frac{2731218467317}{15368042101831} & \frac{9408046702089}{11113171139209} &  0 & 0 & 0 \\
 & -\frac{451086348788}{2902428689909} & -\frac{2682348792572}{7519795681897} &  \frac{12662868775082}{11960479115383} & \frac{3355817975965}{11060851509271} & 0 & 0 \\
 & \frac{647845179188}{3216320057751} &  \frac{73281519250}{8382639484533} &     \frac{552539513391}{3454668386233} &    \frac{3354512671639}{8306763924573} &  \frac{4040}{17871} & 0 \\
 \hline
& \frac{82889}{524892} & 0 & \frac{15625}{83664} & \frac{69875}{102672} &    -\frac{2260}{8211} & \frac14
\end{array}
\\
&
\begin{array}{c |c c c c c c}
 & 0 & 0 & 0 & 0 & 0 & 0 \\
 & \frac{1}{4} & \frac{1}{4} & 0 & 0 & 0 & 0   \\
 & \frac{8611}{62500} & -\frac{1743}{31250} & \frac{1}{4} & 0 & 0 & 0 \\
 & \frac{5012029}{34652500} & -\frac{654441}{2922500} & \frac{174375}{388108} & \frac{1}{4} & 0 & 0 \\
 & \frac{15267082809}{155376265600} & -\frac{71443401}{120774400} & \frac{730878875}{902184768} & \frac{2285395}{8070912} & \frac{1}{4} &        0  \\
 &
\frac{82889}{524892} & 0 & \frac{15625}{83664} & \frac{69875}{102672} &    -\frac{2260}{8211} & \frac14 \\
\hline
& \frac{82889}{524892} & 0 & \frac{15625}{83664} & \frac{69875}{102672} &    -\frac{2260}{8211} & \frac14
\end{array}
\end{aligned}
\end{equation}

\textbf{IMEX-RK5} \cite{KENNEDY2003}
\begin{tiny}
\begin{equation}
\begin{aligned}
&
\begin{array}{c |c c c c c c c c}
 & 0 & 0 & 0 & 0 & 0 & 0 & 0 & 0\\
 & \frac{41}{100} & 0 & 0 & 0 & 0 & 0 & 0 & 0 \\
 & \frac{367902744464}{2072280473677} & \frac{677623207551}{8224143866563}& 0& 0&  0&          0& 0&          0 \\
 &
\frac{1268023523408}{10340822734521} &   0& \frac{1029933939417} {13636558850479} &  0&  0&          0& 0&          0 \\
&
\frac{14463281900351}{6315353703477}&   0& \frac{66114435211212} { 5879490589093} &  -\frac{54053170152839}{4284798021562} &  0 &          0& 0&          0\\
&
\frac{14090043504691} { 34967701212078}&  0&         \frac{15191511035443} { 11219624916014}& -\frac{18461159152457} { 12425892160975}& -\frac{281667163811} { 9011619295870}& 0& 0&          0 \\
&
\frac{19230459214898} { 13134317526959}&  0 &  \frac{21275331358303} { 2942455364971}&  -\frac{38145345988419} { 4862620318723}&  -\frac{1} { 8} &    -\frac{1} { 8} &      0&          0\\
&
\frac{-19977161125411} { 11928030595625}& 0&         -\frac{40795976796054} { 6384907823539}& \frac{177454434618887} { 12078138498510}& \frac{782672205425} { 8267701900261}&  -\frac{69563011059811} { 9646580694205}& \frac{7356628210526} { 4942186776405}& 0 \\
\hline
&
\frac{-872700587467} { 9133579230613}&  0&0& \frac{22348218063261} { 9555858737531}& -\frac{1143369518992} { 8141816002931}& -\frac{39379526789629} { 19018526304540}& \frac{32727382324388} { 42900044865799}& \frac{41} { 200} \\
\end{array}
\\
&
\begin{array}{c |c c c c c c c c}
 &0&0&0& 0&  0&0&  0& 0 \\
 & \frac{41}{200}& \frac{41}{200} & 0& 0&  0&0&  0& 0 \\
 & \frac{41}{400}& -\frac{567603406766} { 11931857230679} & \frac{41} { 200} & 0&  0&0&  0& 0 \\
 & \frac{683785636431}{9252920307686} & 0 & -\frac{110385047103} { 1367015193373} &   \frac{41} { 200}&  0&0&  0& 0 \\
 & \frac{3016520224154} { 10081342136671}& 0& \frac{30586259806659} { 12414158314087} & -\frac{22760509404356} { 11113319521817} & \frac{41} { 200} & 0 &  0& 0 \\
 &
\frac{218866479029} { 1489978393911}&   0& \frac{638256894668} { 5436446318841}&    -\frac{1179710474555} { 5321154724896}&   -\frac{60928119172} { 8023461067671}&   \frac{41} { 200}&  0& 0 \\
&
\frac{1020004230633} { 5715676835656}&  0&\frac{25762820946817} { 25263940353407}& -\frac{2161375909145} { 9755907335909}&   -\frac{211217309593} { 5846859502534}&  -\frac{4269925059573} { 7827059040749}&   \frac{41} { 200} &  0  \\
&
\frac{-872700587467} { 9133579230613}&  0&0& \frac{22348218063261} { 9555858737531}& -\frac{1143369518992} { 8141816002931}& -\frac{39379526789629} { 19018526304540}& \frac{32727382324388} { 42900044865799}& \frac{41} { 200} \\
\hline
&
\frac{-872700587467} { 9133579230613}&  0&0& \frac{22348218063261} { 9555858737531}& -\frac{1143369518992} { 8141816002931}& -\frac{39379526789629} { 19018526304540}& \frac{32727382324388} { 42900044865799}& \frac{41} { 200} \\
\end{array}
\end{aligned}
\end{equation}
\end{tiny}

\newpage
\section{Settings of the experiments}
All the experiments are conducted on NVIDIA A100 GPUs using PyTorch v$2.0.0$ and PETSc v$3.21.0$.
We use a fixed time step for all the methods for a fair comparison since not all the methods are adaptive by design and error estimation incurs additional cost that is scheme specific.
Additional details on data generation for the two dynamics learning problems are provided below.

\subsection{The KS equation}
The equation \eqref{eq:KS} is discretized in space by using a Fourier spectral method and is integrated in time with the exponential time-differencing fourth-order RK method \cite{Kassam2005}.
To obtain the training data, we use a periodic boundary condition and the initial condition
\[
  u = cos(x/22) (1+ sin(x/22))
\]
and collect a time series $\bu(t) \in \mathbb{R}^N$ for a time span of $200$ after discarding the transient states for the initial time period $[0, 1000]$ so that the dynamics is guaranteed to be in the chaotic regime.
In this regime, the solution is extremely sensitive to perturbations in the initial data; for example, perturbations can be amplified by $10^8$ in a time span of $150$ \cite{Kassam2005}. 

\subsection{The viscous Burgers equation}
Solutions of the one-dimensional viscous Burgers equation are represented by $512$ or $1024$ points in state $u$, which are evenly spaced in the spatial domain. We collect $100$ sequential solutions of the viscous Burgers equation over a time interval $[0,5]$ by sampling every $0.1$ time unit for $100$ random initial conditions. In this dataset, $80$ solutions are used as a training dataset, and the remaining 
$20$ solutions are used as a testing dataset.

\newpage
\section{Additional results with GRAND} \label{sec:grand_app}

\begin{figure*}[ht]
\centering
\includegraphics[width=0.96\linewidth]{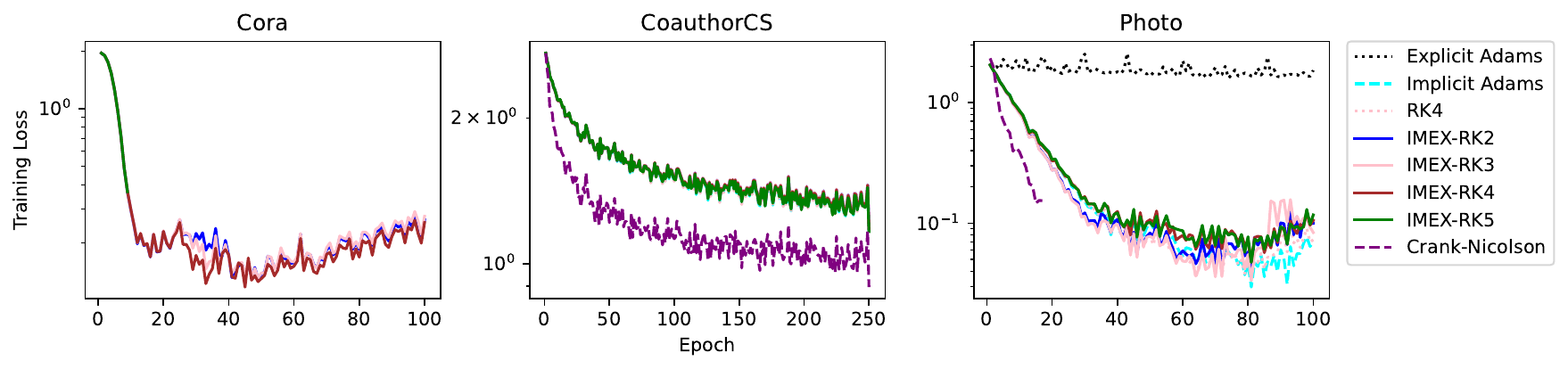}
\vspace{-10pt}
\caption{Training loss vs. epoch for a variety of time integrators using explicit, implicit, and semi-implicit methods. The explicit Adams method cannot complete the training process due to instability. }
\label{fig:grand_loss}
\vspace{-10pt} 
\end{figure*}

Fig. ~\ref{fig:grand_loss} shows that SINODE with IMEX methods converge comparably to neural ODEs with explicit methods, and slightly behind neural ODEs with Crank--Nicolson.

\begin{table}[ht]
\renewcommand{\arraystretch}{1.2} 
    \begin{center}
        \caption{Forward (before $+$) and backward (after $+$) NFEs per epoch for neural ODEs with explicit methods and implicit methods and SINODE with IMEX methods. The numbers are rounded to the closest integer.}
        \label{tab:grand_nfe_detailed}
        \small
        \begin{tabular}{c  c  c  c} 
           \toprule
            & \multicolumn{3}{c}{NFEs} \\
            \cline{2-4}
            & Cora & CoauthorCS & Photo \\
          \midrule
RK4 & 29272+14636 & 5008+2504 & 5736+2836 \\
Explicit Adams & 7330+3665 & 1264+632 & 1446+723 \\
Implicit Adams & 302+1550 & 44+29 & 40+22 \\
IMEX-RK2 & \textbf{76+38} & \textbf{16+8} & \textbf{16+8} \\
IMEX-RK3 & \textbf{152+76} & \textbf{32+16} & \textbf{32+16} \\
IMEX-RK4 & \textbf{228+114} & \textbf{48+24} & \textbf{48+24} \\
IMEX-RK5 & \textbf{304+152} & \textbf{64+32} & \textbf{64+32} \\
Crank-Nicolson & 520+91 & 149+25 & 419+28 \\
\bottomrule
        \end{tabular}
    \end{center} 
\end{table}

\newpage
\section{Additional results with KSE} \label{sec:kse_app}

\begin{figure}[!ht]
  \centering
  \includegraphics[width=0.7\linewidth]{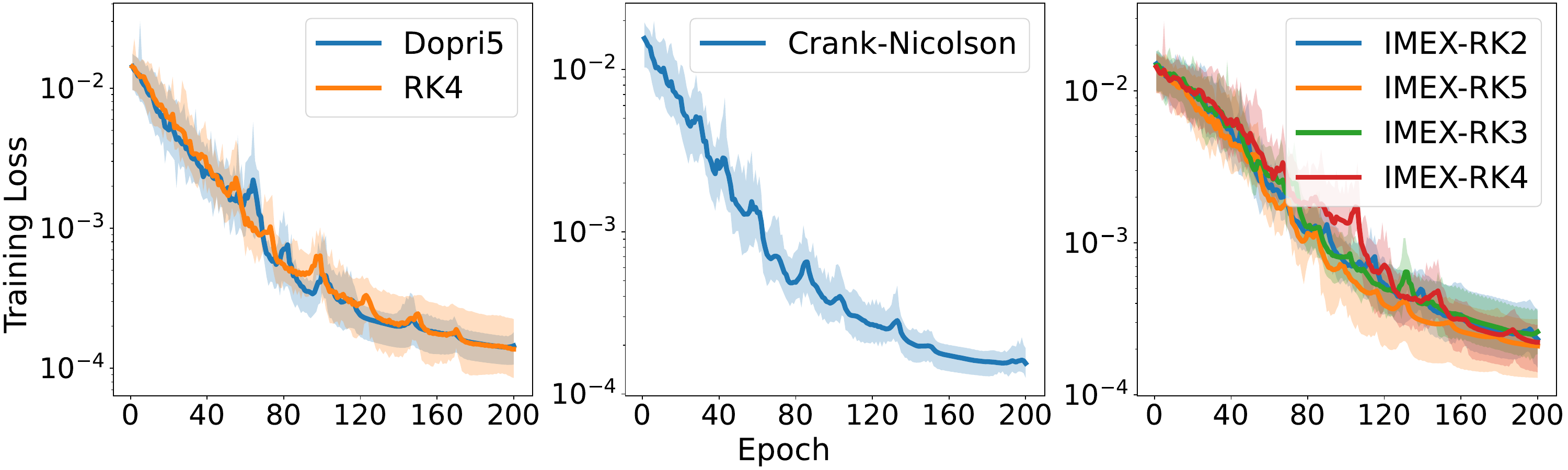}
  \caption{Training loss for KSE on a uniform grid of size $64$. 
  }
  \label{fig:kse64_conv}
\end{figure}

Fig.~\ref{fig:kse64_conv} shows that SINODE leads to a stable training algorithm that converges comparably to the explicit methods and the fully implicit method.
Higher-order IMEX methods such as the 5th-order IMEX-RK tend to exhibit slightly better convergence because of their higher accuracy.
\begin{figure}[!ht]
  \centering
  \includegraphics[width=0.3\linewidth]{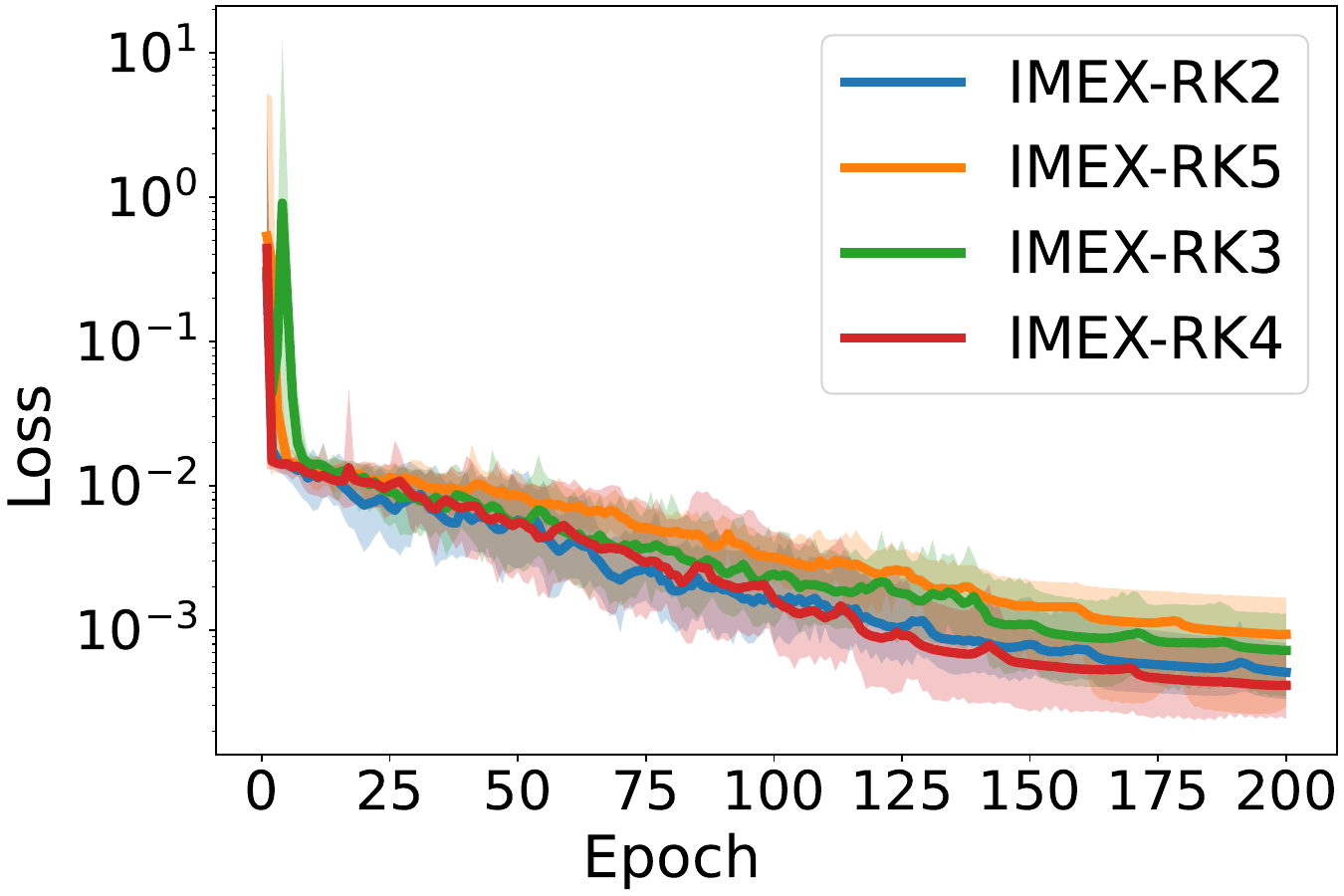}
  \caption{Training loss for KSE on a uniform grid of size $512$. 
  }
  \label{fig:kse512_conv}
\end{figure}
 \begin{table}[!ht]
\renewcommand{\arraystretch}{1.2} 
    \begin{center}
        \caption{Architecture of the NN used in Sec.~\ref{sec:kse} for the KSE. The grid size used to generate the training data is $64$. “Shape” indicates the dimension of each layer, “Activation” is the corresponding activation functions, and “Initial Weights" is the distribution used to initialize the weights of the NNs.}
        \small
        \label{tab:kse_nn64}
        \begin{tabular}{c  c  c  c}
        \toprule
        Function & Shape & Activation & Initial Weights \\
        $\bG$ & 64/200/200/200/200/64 & ReLU/ReLU/ReLU/ReLU/Linear &  $\N$ (0,0.01) \\
        $\J$ &  64/64 &  Linear & $-$ \\
       \bottomrule
        \end{tabular}
    \end{center}
    \vspace{-10pt}
\end{table}
 \begin{table}[h!]
\renewcommand{\arraystretch}{1.2} 
    \begin{center}
        \caption{Architecture of the NN used in Sec.~\ref{sec:kse} for the KSE. The grid size used to generate the training data is $512$. “Shape” indicates the dimension of each layer, “Activation” is the corresponding activation functions,  and “Initial Weights" is the distribution used to initialize the weights of the NNs.}
        \small
        \label{tab:kse_nn512}
        \begin{tabular}{c  c  c  c}
        \toprule
        Function & Shape & Activation & Initial Weights \\
        $\bG$ & 512/1600/1600/1600/1600/512 & ReLU/ReLU/ReLU/ReLU/Linear &  $\N$ (0,0.01) \\
        $\J$ &  512/512 &  Linear & $-$ \\
       \bottomrule
        \end{tabular}
    \end{center} 
\end{table}

\newpage

\section{Additional results with Burgers} \label{sec:burgers_app}

\begin{figure}[!ht]
\centering
  \includegraphics[width=0.65\linewidth]{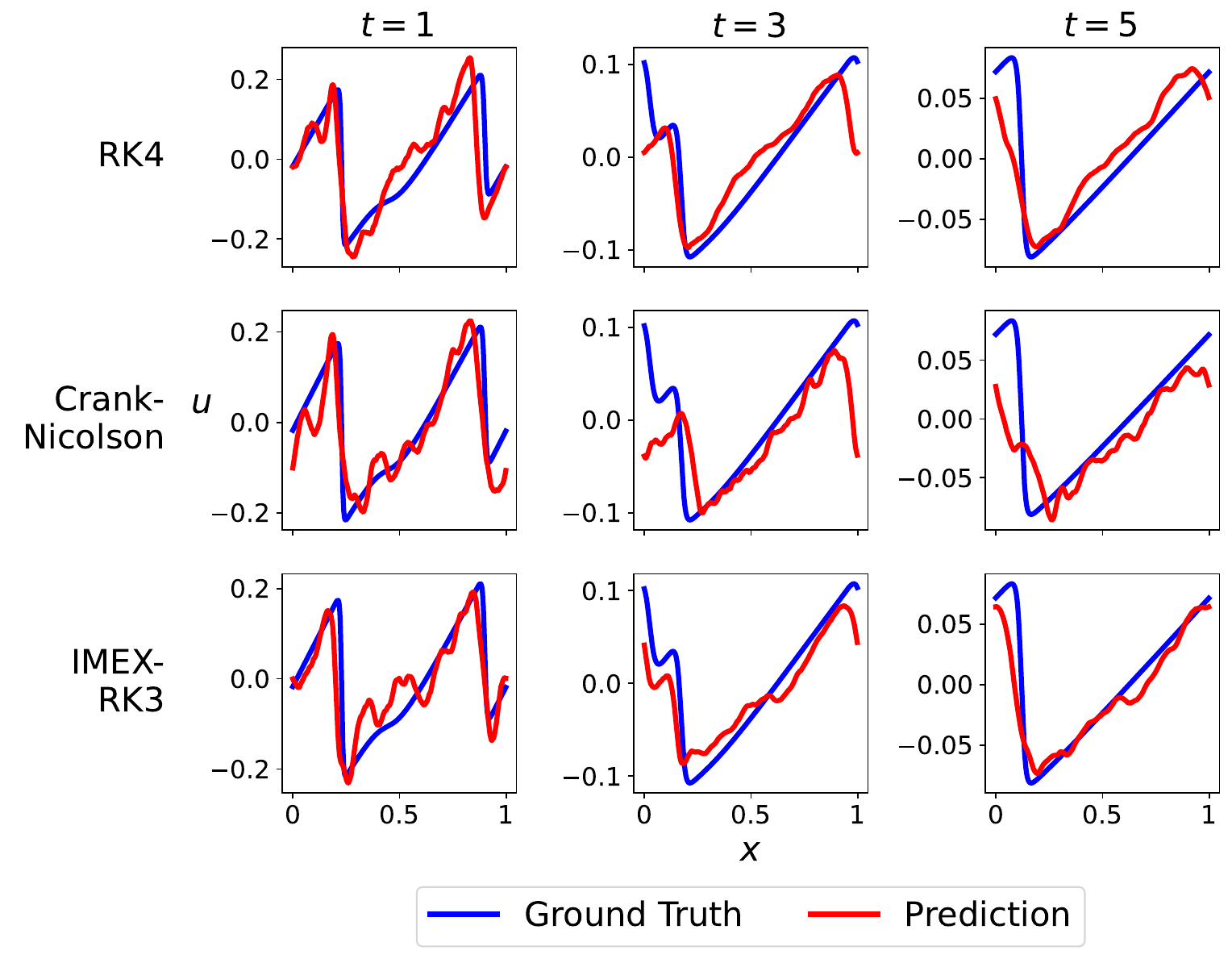}
\caption{Snapshots of VBE trajectories of ground truth and model predictions for grid size $512$ from different methods (rows) at different times (columns). 
}
\label{fig:burgers_snapshot_512}
\end{figure}

\begin{figure}[!ht]
\centering
  \includegraphics[width=0.65\linewidth]{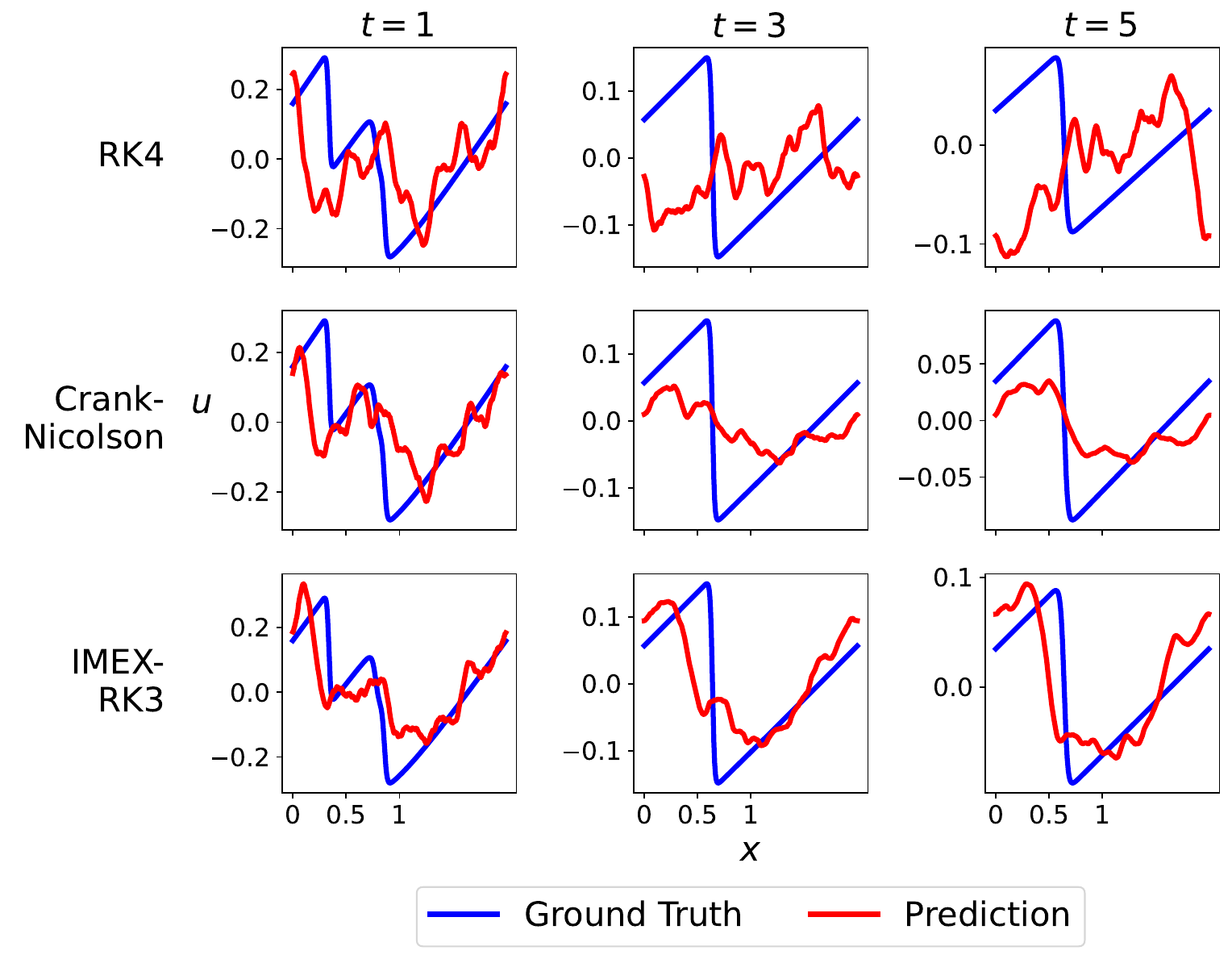}
\caption{Snapshots of VBE trajectories of ground truth and model predictions for grid size $1024$ from different models. The rows represent RK4, Crank--Nicolson, and IMEX-RK3, respectively. The columns indicate solutions taken at $t=1$, $3$, and $5$, respectively. }
\label{fig:burgers_snapshot_1024}
\end{figure}

In Fig.~\ref{fig:burgers_snapshot_512} and Fig.~\ref{fig:burgers_snapshot_1024} we compare prediction performance among RK4, Crank--Nicolson, and IMEX-RK3 on one solution in the testing dataset. IMEX-RK3 effectively captures the evolution of turbulence shape as time advances.

 \begin{table}[!ht]
\renewcommand{\arraystretch}{1.2} 
    \begin{center}
        \caption{Architecture of the NN used in Sec.~\ref{sec:burgers} for Burgers. The grid size used to generate the training data is $512$. “Shape” indicates the dimension of each layer, “Activation” is the corresponding activation functions, and “Initial Weights" is the distribution used to initialize the weights of the NNs.}
        \small
        \label{tab:burgers_nn512}
        \begin{tabular}{c  c  c  c}
        \toprule
        Function & Shape & Activation & Initial Weights \\
        $\bG$ & 512/576/576/576/576/512 & ReLU/ReLU/ReLU/ReLU/Linear &  $\N$ (0,0.1) \\
        $\J$ &  512/512 &  Linear & $-$ \\
       \bottomrule
        \end{tabular}
    \end{center} 
\end{table}

 \begin{table}[!ht]
\renewcommand{\arraystretch}{1.2} 
    \begin{center}
        \caption{Architecture of the NN used in Sec.~\ref{sec:kse} for Burgers. The grid size used to generate the training data is $1024$. “Shape” indicates the dimension of each layer, “Activation” the corresponding activation functions, “Initial Weights" is the distribution used to initialize the weights of the NNs.}
        \small
        \label{tab:burgers_nn1024}
        \begin{tabular}{c  c  c  c}
        \toprule
        Function & Shape & Activation & Initial Weights \\
        $\bG$ & 1024/1152/1152/1152/1152/1024 & ReLU/ReLU/ReLU/ReLU/Linear &  $\N$ (0,0.1) \\
        $\J$ &  1024/1024 &  Linear & $-$ \\
       \bottomrule
        \end{tabular}
    \end{center} 
\end{table}

\begin{figure}[!ht]
  \centering
  \includegraphics[width=0.75\linewidth]{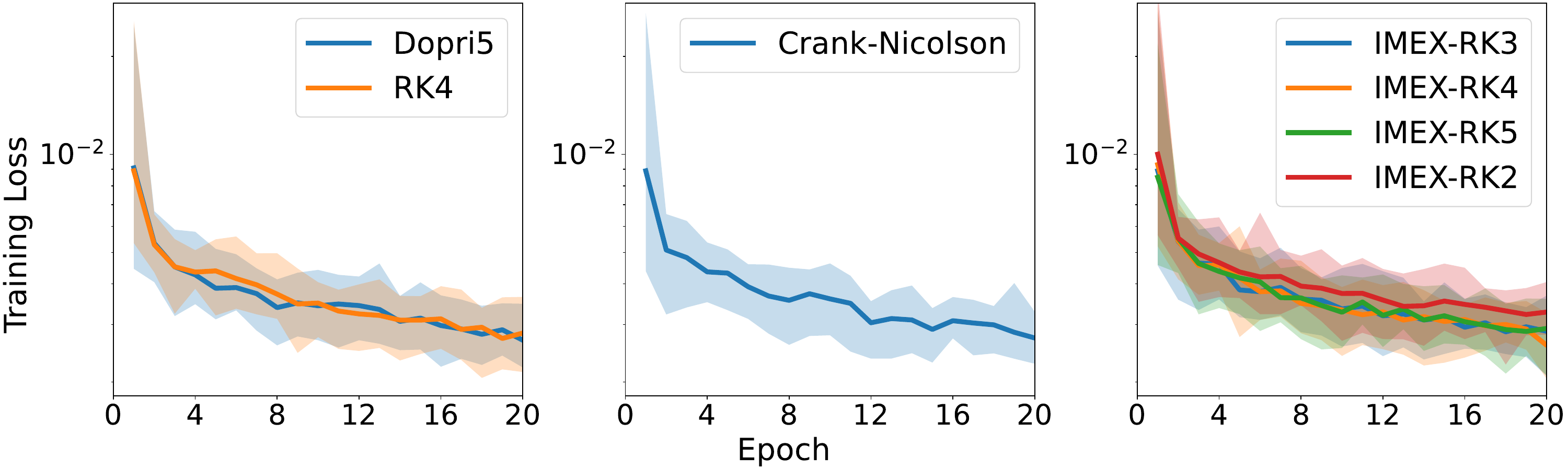}
  \caption{Training loss for the Burgers equation on a uniform grid of size $512$. 
  }
  \label{fig:burgers_conv512}
\end{figure}

\begin{figure}[!ht]
  \centering
  \includegraphics[width=0.75\linewidth]{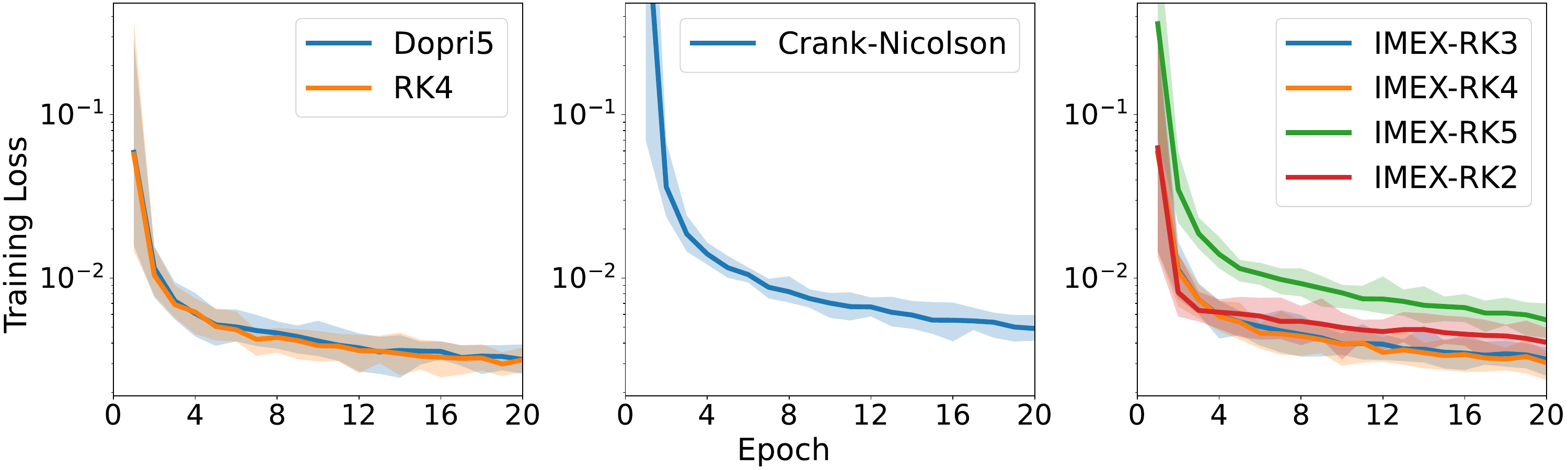}
  \caption{Training loss for the Burgers equation on a uniform grid of size $1024$. 
  }
  \label{fig:burgers_conv1024}
\end{figure}



\end{document}

%% file: preample_shared.tex
\usepackage{bm}

\newcommand{\bu}{\bm{u}}
\newcommand{\bU}{\bm{U}}

\newcommand{\bp}{\bm{p}}

\newcommand{\R}{\mathbb{R}}
\newcommand{\I}{\bm{\mathcal{I}}}

\newcommand{\J}{\bm{\mathcal{J}}}

\newcommand{\N}{\bm{\mathcal{N}}}

\newcommand{\blambda}{\bm{\lambda}}

\newcommand{\bmu}{\bm{\mu}}
\newcommand{\f}{\bm{f}}

\newcommand{\fp}{\f_{\bp}}
\newcommand{\fu}{\f_{\bu}}

\newcommand{\bG}{\bm{\mathcal{G}}}
\newcommand{\bGu}{\bG_{\bu}}
\newcommand{\bGp}{\bG_{\bp}}
\newcommand{\bH}{\bm{\mathcal{H}}}
\newcommand{\bHu}{\bH_{\bu}}
\newcommand{\bHp}{\bH_{\bp}}

%% file: main.bbl
\begin{thebibliography}{35}
\providecommand{\natexlab}[1]{#1}

\bibitem[{Ascher, Ruuth, and Spiteri(1997)}]{ascher1997implicit}
Ascher, U.~M.; Ruuth, S.~J.; and Spiteri, R.~J. 1997.
\newblock Implicit-explicit {Runge--Kutta} methods for time-dependent partial
  differential equations.
\newblock \emph{Applied Numerical Mathematics}, 25(2-3): 151--167.

\bibitem[{Bai, Kolter, and Koltun(2019)}]{Bai2019}
Bai, S.; Kolter, J.~Z.; and Koltun, V. 2019.
\newblock Deep equilibrium models.
\newblock In \emph{Advances in Neural Information Processing Systems},
  volume~32.

\bibitem[{Baker et~al.(2022)Baker, Xia, Wang, Cherkaev, Narayan, Chen, Xin,
  Bertozzi, Osher, and Wang}]{baker2022proximal}
Baker, J.; Xia, H.; Wang, Y.; Cherkaev, E.; Narayan, A.; Chen, L.; Xin, J.;
  Bertozzi, A.~L.; Osher, S.~J.; and Wang, B. 2022.
\newblock Proximal Implicit {ODE} Solvers for Accelerating Learning Neural
  {ODEs}.
\newblock arXiv:2204.08621.

\bibitem[{Balay et~al.(2023)Balay, Abhyankar, Adams, Benson, Brown, Brune,
  Buschelman, Constantinescu, Dalcin, Dener, Eijkhout, Faibussowitsch, Gropp,
  Hapla, Isaac, Jolivet, Karpeev, Kaushik, Knepley, Kong, Kruger, May, McInnes,
  Mills, Mitchell, Munson, Roman, Rupp, Sanan, Sarich, Smith, Zampini, Zhang,
  Zhang, and Zhang}]{petsc-user-ref}
Balay, S.; Abhyankar, S.; Adams, M.~F.; Benson, S.; Brown, J.; Brune, P.;
  Buschelman, K.; Constantinescu, E.; Dalcin, L.; Dener, A.; Eijkhout, V.;
  Faibussowitsch, J.; Gropp, W.~D.; Hapla, V.; Isaac, T.; Jolivet, P.; Karpeev,
  D.; Kaushik, D.; Knepley, M.~G.; Kong, F.; Kruger, S.; May, D.~A.; McInnes,
  L.~C.; Mills, R.~T.; Mitchell, L.; Munson, T.; Roman, J.~E.; Rupp, K.; Sanan,
  P.; Sarich, J.; Smith, B.~F.; Zampini, S.; Zhang, H.; Zhang, H.; and Zhang,
  J. 2023.
\newblock {PETSc/TAO} Users Manual.
\newblock Technical Report ANL-21/39 -- Revision 3.19, Argonne National
  Laboratory.

\bibitem[{Boscarino(2007)}]{boscarino2007error}
Boscarino, S. 2007.
\newblock Error analysis of {IMEX Runge--Kutta} methods derived from
  differential-algebraic systems.
\newblock \emph{SIAM Journal on Numerical Analysis}, 45(4): 1600--1621.

\bibitem[{Boscarino and Russo(2009)}]{boscarino2009class}
Boscarino, S.; and Russo, G. 2009.
\newblock On a class of uniformly accurate {IMEX Runge--Kutta} schemes and
  applications to hyperbolic systems with relaxation.
\newblock \emph{SIAM Journal on Scientific Computing}, 31(3): 1926--1945.

\bibitem[{Chamberlain et~al.(2021)Chamberlain, Rowbottom, Gorinova, Bronstein,
  Webb, and Rossi}]{chamberlain2021grand}
Chamberlain, B.; Rowbottom, J.; Gorinova, M.~I.; Bronstein, M.; Webb, S.; and
  Rossi, E. 2021.
\newblock {Grand: Graph neural diffusion}.
\newblock In \emph{International Conference on Machine Learning}, 1407--1418.
  PMLR.

\bibitem[{Chen et~al.(2018)Chen, Rubanova, Bettencourt, Duvenaud, Chen,
  Rubanova, Bettencourt, and Duvenaud}]{Chen2018}
Chen, T.~Q.; Rubanova, Y.; Bettencourt, J.; Duvenaud, D.; Chen, R.~T.;
  Rubanova, Y.; Bettencourt, J.; and Duvenaud, D. 2018.
\newblock Neural Ordinary Differential Equations.
\newblock In \emph{Advances in Neural Information Processing Systems}.

\bibitem[{Daulbaev et~al.(2020)Daulbaev, Katrutsa, Markeeva, Gusak, Cichocki,
  and Oseledets}]{Daulbaev2020}
Daulbaev, T.; Katrutsa, A.; Markeeva, L.; Gusak, J.; Cichocki, A.; and
  Oseledets, I. 2020.
\newblock {Interpolation Technique to Speed Up Gradients Propagation in Neural
  ODEs}.
\newblock In Larochelle, H.; Ranzato, M.; Hadsell, R.; Balcan, M.; and Lin, H.,
  eds., \emph{Advances in Neural Information Processing Systems}, volume~33,
  16689--16700. Curran Associates, Inc.

\bibitem[{Dupont, Doucet, and Teh(2019)}]{Dupont2019}
Dupont, E.; Doucet, A.; and Teh, Y.~W. 2019.
\newblock {Augmented Neural ODEs}.
\newblock In Wallach, H.; Larochelle, H.; Beygelzimer, A.; d\textquotesingle
  Alch\'{e}-Buc, F.; Fox, E.; and Garnett, R., eds., \emph{Advances in Neural
  Information Processing Systems}, volume~32. Curran Associates, Inc.

\bibitem[{Fung et~al.(2022)Fung, Heaton, Li, McKenzie, Osher, and
  Yin}]{Fung2022}
Fung, S.~W.; Heaton, H.; Li, Q.; McKenzie, D.; Osher, S.; and Yin, W. 2022.
\newblock {JFB: Jacobian}-Free Backpropagation for Implicit Networks.
\newblock In \emph{Proceedings of the 36th AAAI Conference on Artificial
  Intelligence, AAAI 2022}, volume~36.

\bibitem[{Gholaminejad, Keutzer, and Biros(2019)}]{ANODE}
Gholaminejad, A.; Keutzer, K.; and Biros, G. 2019.
\newblock {ANODE}: Unconditionally Accurate Memory-Efficient Gradients for
  Neural {ODEs}.
\newblock In \emph{Proceedings of the Twenty-Eighth International Joint
  Conference on Artificial Intelligence, {IJCAI-19}}, 730--736. International
  Joint Conferences on Artificial Intelligence Organization.

\bibitem[{Haber et~al.(2019)Haber, Lensink, Treister, and Ruthotto}]{haber19a}
Haber, E.; Lensink, K.; Treister, E.; and Ruthotto, L. 2019.
\newblock {IMEX}net A Forward Stable Deep Neural Network.
\newblock In Chaudhuri, K.; and Salakhutdinov, R., eds., \emph{Proceedings of
  the 36th International Conference on Machine Learning}, volume~97 of
  \emph{Proceedings of Machine Learning Research}, 2525--2534. PMLR.

\bibitem[{Kassam and Trefethen(2005)}]{Kassam2005}
Kassam, A.-K.; and Trefethen, L.~N. 2005.
\newblock Fourth-Order Time-Stepping for Stiff {PDEs}.
\newblock \emph{SIAM Journal on Scientific Computing}, 26(4): 1214--1233.

\bibitem[{Kennedy and Carpenter(2003{\natexlab{a}})}]{kennedy2003additive}
Kennedy, C.~A.; and Carpenter, M.~H. 2003{\natexlab{a}}.
\newblock Additive {Runge--Kutta} schemes for convection--diffusion--reaction
  equations.
\newblock \emph{Applied Numerical Mathematics}, 44(1-2): 139--181.

\bibitem[{Kennedy and Carpenter(2003{\natexlab{b}})}]{KENNEDY2003}
Kennedy, C.~A.; and Carpenter, M.~H. 2003{\natexlab{b}}.
\newblock Additive {Runge--Kutta} schemes for convection–diffusion–reaction
  equations.
\newblock \emph{Applied Numerical Mathematics}, 44(1): 139--181.

\bibitem[{Kidger(2022)}]{kidger2022neural}
Kidger, P. 2022.
\newblock On Neural Differential Equations.
\newblock \emph{arXiv preprint arXiv:2202.02435}.

\bibitem[{Kidger, Chen, and Lyons(2020)}]{Kidger2020HeyTN}
Kidger, P.; Chen, R. T.~Q.; and Lyons, T. 2020.
\newblock "Hey, that's not an ODE": Faster ODE Adjoints via Seminorms.
\newblock In \emph{International Conference on Machine Learning}.

\bibitem[{Kim et~al.(2021)Kim, Ji, Deng, Ma, and Rackauckas}]{kim2021stiff}
Kim, S.; Ji, W.; Deng, S.; Ma, Y.; and Rackauckas, C. 2021.
\newblock {Stiff neural ordinary differential equations}.
\newblock \emph{Chaos: An Interdisciplinary Journal of Nonlinear Science},
  31(9).

\bibitem[{Linot et~al.(2022)Linot, Burby, Tang, Balaprakash, Graham, and
  Maulik}]{Linot2022}
Linot, A.~J.; Burby, J.~W.; Tang, Q.; Balaprakash, P.; Graham, M.~D.; and
  Maulik, R. 2022.
\newblock Stabilized Neural Ordinary Differential Equations for Long-Time
  Forecasting of Dynamical Systems.
\newblock \emph{Journal of Computational Physics}, 474.

\bibitem[{Massaroli et~al.(2020)Massaroli, Poli, Park, Yamashita, and
  Asama}]{Massaroli2020}
Massaroli, S.; Poli, M.; Park, J.; Yamashita, A.; and Asama, H. 2020.
\newblock {Dissecting Neural ODEs}.
\newblock In Larochelle, H.; Ranzato, M.; Hadsell, R.; Balcan, M.; and Lin, H.,
  eds., \emph{Advances in Neural Information Processing Systems}, volume~33,
  3952--3963. Curran Associates, Inc.

\bibitem[{Matsubara, Miyatake, and Yaguchi(2023)}]{Matsubara2023}
Matsubara, T.; Miyatake, Y.; and Yaguchi, T. 2023.
\newblock The Symplectic Adjoint Method: Memory-Efficient Backpropagation of
  Neural-Network-Based Differential Equations.
\newblock \emph{IEEE Transactions on Neural Networks and Learning Systems},
  1--13.

\bibitem[{Mills et~al.(2021)Mills, Adams, Balay, Brown, Dener, Knepley, Kruger,
  Morgan, Munson, Rupp, Smith, Zampini, Zhang, and Zhang}]{Mills2021}
Mills, R.~T.; Adams, M.~F.; Balay, S.; Brown, J.; Dener, A.; Knepley, M.;
  Kruger, S.~E.; Morgan, H.; Munson, T.; Rupp, K.; Smith, B.~F.; Zampini, S.;
  Zhang, H.; and Zhang, J. 2021.
\newblock {Toward performance-portable PETSc for GPU-based exascale systems}.
\newblock \emph{Parallel Computing}, 108: 102831.

\bibitem[{Onken and Ruthotto(2020)}]{onken2020discretizeoptimize}
Onken, D.; and Ruthotto, L. 2020.
\newblock Discretize-Optimize vs. Optimize-Discretize for Time-Series
  Regression and Continuous Normalizing Flows.
\newblock \emph{arXiv preprint arXiv:2005.13420}.

\bibitem[{Pareschi and Russo(2005)}]{pareschi2005implicit}
Pareschi, L.; and Russo, G. 2005.
\newblock Implicit--explicit {Runge--Kutta} schemes and applications to
  hyperbolic systems with relaxation.
\newblock \emph{Journal of Scientific Computing}, 25: 129--155.

\bibitem[{Poli et~al.(2019)Poli, Massaroli, Park, Yamashita, Asama, and
  Park}]{poli2019graph}
Poli, M.; Massaroli, S.; Park, J.; Yamashita, A.; Asama, H.; and Park, J. 2019.
\newblock Graph Neural Ordinary Differential Equations.
\newblock \emph{arXiv preprint arXiv:1911.07532}.

\bibitem[{Winston and Kolter(2020)}]{Winston2020}
Winston, E.; and Kolter, J.~Z. 2020.
\newblock Monotone operator equilibrium networks.
\newblock In \emph{Advances in Neural Information Processing Systems}, volume
  2020-December.

\bibitem[{Xhonneux, Qu, and Tang(2020)}]{Xhonneux2020}
Xhonneux, L.-P.; Qu, M.; and Tang, J. 2020.
\newblock Continuous Graph Neural Networks.
\newblock In III, H.~D.; and Singh, A., eds., \emph{Proceedings of the 37th
  International Conference on Machine Learning}, volume 119 of
  \emph{Proceedings of Machine Learning Research}, 10432--10441. PMLR.

\bibitem[{Zhang and Constantinescu(2021)}]{Zhang2021iccs}
Zhang, H.; and Constantinescu, E.~M. 2021.
\newblock {Revolve-based adjoint checkpointing for multistage time
  integration}.
\newblock In \emph{{International Conference on Computational Science}}.
  (online), in main track.

\bibitem[{Zhang and Constantinescu(2023)}]{Zhang2023JCS}
Zhang, H.; and Constantinescu, E.~M. 2023.
\newblock Optimal checkpointing for adjoint multistage time-stepping schemes.
\newblock \emph{Journal of Computational Science}, 66: 101913.

\bibitem[{Zhang and Sandu(2014)}]{Zhang2014}
Zhang, H.; and Sandu, A. 2014.
\newblock {FATODE: a library for forward, adjoint, and tangent linear
  integration of ODEs}.
\newblock \emph{SIAM Journal on Scientific Computing}, 36(5): C504--C523.

\bibitem[{Zhang and Zhao(2022)}]{Zhang2022pnode}
Zhang, H.; and Zhao, W. 2022.
\newblock A Memory-Efficient Neural Ordinary Differential Equation Framework
  Based on High-Level Adjoint Differentiation.
\newblock \emph{IEEE Transactions on Artificial Intelligence}, 1--11.

\bibitem[{Zhang et~al.(2019)Zhang, Yao, Gholami, Gonzalez, Keutzer, Mahoney,
  and Biros}]{ANODEV2}
Zhang, T.; Yao, Z.; Gholami, A.; Gonzalez, J.~E.; Keutzer, K.; Mahoney, M.~W.;
  and Biros, G. 2019.
\newblock {ANODEV2}: A Coupled Neural {ODE} Framework.
\newblock In \emph{Advances in Neural Information Processing Systems},
  volume~32. Curran Associates, Inc.

\bibitem[{Zhuang et~al.(2020)Zhuang, Dvornek, Li, Tatikonda, Papademetris, and
  Duncan}]{zhuang2020}
Zhuang, J.; Dvornek, N.; Li, X.; Tatikonda, S.; Papademetris, X.; and Duncan,
  J. 2020.
\newblock Adaptive Checkpoint Adjoint Method for Gradient Estimation in Neural
  {ODE}.
\newblock In \emph{Proceedings of the 37th International Conference on Machine
  Learning}, volume 119, 11639--11649.

\bibitem[{Zhuang et~al.(2021)Zhuang, Dvornek, Tatikond, and
  Duncan}]{zhuang2021mali}
Zhuang, J.; Dvornek, N.~C.; Tatikond, S.; and Duncan, J.~S. 2021.
\newblock {MALI}: A memory efficient and reverse accurate integrator for Neural
  {ODE}s.
\newblock In \emph{International Conference on Learning Representations}.

\end{thebibliography}
